\documentclass[10pt,twocolumn,letterpaper]{article}

\usepackage[pagenumbers]{cvpr} 

\makeatletter
\@namedef{ver@everyshi.sty}{}
\makeatother 

\usepackage{booktabs}

\usepackage{mathptmx}
\usepackage{graphicx}
\usepackage{amsmath}
\usepackage{amsthm}
\usepackage{amssymb} 
\usepackage{pifont}
\newcommand{\cmark}{\ding{51}}%
\newcommand{\xmark}{\ding{55}}%

\usepackage{wrapfig}

\usepackage[pagebackref=true,breaklinks=true,colorlinks,bookmarks=false]{hyperref}

\usepackage[capitalize]{cleveref}
\crefname{section}{Sec.}{Secs.}
\Crefname{section}{Section}{Sections}
\Crefname{table}{Table}{Tables}
\crefname{table}{Tab.}{Tabs.}

\usepackage{xcolor}
\usepackage{tikz}
\usepackage{changepage}
\usepackage{tikz-3dplot}
\usepackage{tikzscale}
\usetikzlibrary{plotmarks}
\usepackage{pgfplots}
\usetikzlibrary{tikzmark,calc} 
\usetikzlibrary{decorations.pathreplacing, decorations.markings}
  
\usepackage[capbesideposition=right]{floatrow}

\xdefinecolor{navyblue}{RGB}{19 41 75}
\xdefinecolor{carolinablue}{RGB}{123 175 212}
\xdefinecolor{cb}{RGB}{123 175 212}
\xdefinecolor{flatgrey}{RGB}{162 170 173}
\xdefinecolor{matcha}{RGB}{183 224 176}
\xdefinecolor{turquoise}{RGB}{23, 165, 137}
\xdefinecolor{creamblue}{RGB}{52, 152, 219}

\xdefinecolor{palegreen}{RGB}{152, 251, 152}
\xdefinecolor{lightblue}{RGB}{135, 206, 235}
\xdefinecolor{plum}{RGB}{215, 181, 216}

\xdefinecolor{myyellow}{RGB}{255, 239, 159}
\xdefinecolor{myblue}{RGB}{181, 222, 255}
\xdefinecolor{mypurple}{RGB}{202, 184, 255}

\newcommand{\dashedLayer}[6]{
			\def\a{#1} 
			\def\b{0.02}
			\def\c{#2} 
			\def\t{#3} 
			\def\d{#4} 

			\draw[line width=0.15mm, dash pattern=on 2.25pt off 1.8pt](\c+\t,0,\d) -- (\c+\t,\a,\d) -- (\t,\a,\d);                                                      
			\draw[line width=0.15mm, dash pattern=on 2.25pt off 1.8pt](\t,0,\a+\d) -- (\c+\t,0,\a+\d) node[midway,below] {#6} -- (\c+\t,\a,\a+\d) -- (\t,\a,\a+\d) -- (\t,0,\a+\d); 
			\draw[line width=0.15mm, dash pattern=on 2.25pt off 1.8pt](\c+\t,0,\d) -- (\c+\t,0,\a+\d);
			\draw[line width=0.15mm, dash pattern=on 2.25pt off 1.8pt](\c+\t,\a,\d) -- (\c+\t,\a,\a+\d);
			\draw[line width=0.15mm, dash pattern=on 2.25pt off 1.8pt](\t,\a,\d) -- (\t,\a,\a+\d);
			
			\draw[line width=0.15mm] (\c+\t,0,\d) -- (\c+\t,\a,\d);
			\draw[line width=0.15mm] (\c+\t,0,\d) -- (\c+\t,0,\a+\d);
			\draw[line width=0.15mm] (\c+\t,\a,\d) -- (\c+\t,\a,\a+\d);
			\draw[line width=0.15mm] (\c+\t,0,\a+\d) -- (\c+\t,\a,\a+\d);
			
			\filldraw[#5] (\t+\b,\b,\a+\d) -- (\c+\t-\b,\b,\a+\d) -- (\c+\t-\b,\a-\b,\a+\d) -- (\t+\b,\a-\b,\a+\d) -- (\t+\b,\b,\a+\d); 
			\filldraw[#5] (\t+\b,\a,\a-\b+\d) -- (\c+\t-\b,\a,\a-\b+\d) -- (\c+\t-\b,\a,\b+\d) -- (\t+\b,\a,\b+\d);
			\ifthenelse {\equal{#5} {}}
			{} 
			{\filldraw[#5] (\c+\t,\b,\a-\b+\d) -- (\c+\t,\b,\b+\d) -- (\c+\t,\a-\b,\b+\d) -- (\c+\t,\a-\b,\a-\b+\d);} 
		}

\newcommand{\networkLayer}[6]{
			\def\a{#1} 
			\def\b{0.02}
			\def\c{#2} 
			\def\t{#3} 
			\def\d{#4} 

			\draw[line width=0.15mm](\c+\t,0,\d) -- (\c+\t,\a,\d) -- (\t,\a,\d);                                                      
			\draw[line width=0.15mm](\t,0,\a+\d) -- (\c+\t,0,\a+\d) node[midway,below] {#6} -- (\c+\t,\a,\a+\d) -- (\t,\a,\a+\d) -- (\t,0,\a+\d); 
			\draw[line width=0.15mm](\c+\t,0,\d) -- (\c+\t,0,\a+\d);
			\draw[line width=0.15mm](\c+\t,\a,\d) -- (\c+\t,\a,\a+\d);
			\draw[line width=0.15mm](\t,\a,\d) -- (\t,\a,\a+\d);

			\filldraw[#5] (\t+\b,\b,\a+\d) -- (\c+\t-\b,\b,\a+\d) -- (\c+\t-\b,\a-\b,\a+\d) -- (\t+\b,\a-\b,\a+\d) -- (\t+\b,\b,\a+\d); 
			\filldraw[#5] (\t+\b,\a,\a-\b+\d) -- (\c+\t-\b,\a,\a-\b+\d) -- (\c+\t-\b,\a,\b+\d) -- (\t+\b,\a,\b+\d);
			\ifthenelse {\equal{#5} {}}
			{} 
			{\filldraw[#5] (\c+\t,\b,\a-\b+\d) -- (\c+\t,\b,\b+\d) -- (\c+\t,\a-\b,\b+\d) -- (\c+\t,\a-\b,\a-\b+\d);} 
		}
		
\usepackage[linesnumbered,ruled,vlined]{algorithm2e}

\SetCommentSty{mycommfont}
\SetKwInput{Input}{Input}                
\SetKwInput{Output}{Output}         
\SetKwInput{Dataset}{Dataset}     
\SetKwInput{Settings}{Settings}      
\SetKwInput{Initialization}{Initialization}      
\SetAlgorithmName{Alg.}{Alg.}{List of Algorithms} 
\makeatletter
\newcommand{\algorithmfootnote}[2][\footnotesize]{%
  \let\old@algocf@finish\@algocf@finish
  \def\@algocf@finish{\old@algocf@finish
    \leavevmode\rlap{\begin{minipage}{\linewidth}
    #1#2
    \end{minipage}}%
  }%
}
\makeatother

\usepackage{floatrow}

\usepackage{caption} 
\usepackage{subcaption}
\usepackage{makecell}
\usepackage{booktabs}
\usepackage{multirow}

\newcolumntype{P}[1]{>{\centering\arraybackslash}p{#1}}

\usepackage{hyperref} 
  
\newtheoremstyle{bfnote}%
  {}{}
  {\itshape}{}
  {\bfseries}{.}
  { }{\thmname{#1}\thmnumber{ #2}\thmnote{ (#3)}}
\theoremstyle{bfnote}

\newtheorem*{notation*}{Notation}

\usepackage{url}

\pgfplotsset{compat=1.18}
\begin{document}

\title{Unifying Tracking and Image-Video Object Detection}
\author{Peirong Liu\textsuperscript{1} \quad Rui Wang\textsuperscript{2} \quad Pengchuan Zhang\textsuperscript{2} \quad Omid Poursaeed\textsuperscript{2} \quad
Yipin Zhou\textsuperscript{2}  \\ Xuefei Cao\textsuperscript{2} \quad Sreya Dutta Roy\textsuperscript{2} \quad Ashish Shah\textsuperscript{2} \quad Ser-Nam Lim\textsuperscript{2} \vspace{0.3cm} \\ 
\textsuperscript{1}UNC-Chapel Hill \quad \textsuperscript{2}Meta AI 
} 


\twocolumn[{
\maketitle

\begin{center}
\captionsetup{type=figure}
\resizebox{1.0\linewidth}{!}{
	\begin{tikzpicture}
		\tikzstyle{myarrows}=[line width=0.6mm,draw=blue!50,-triangle 45,postaction={draw, line width=0.05mm, shorten >=0.01mm, -}]
		\tikzstyle{mylines}=[line width=0.6mm]

		\pgfmathsetmacro{\dx}{3.65} 
		\pgfmathsetmacro{\dy}{2}
		\pgfmathsetmacro{\sx}{0.2} 
		\pgfmathsetmacro{\sy}{-0.9}

		\draw[dashed, color = cb, line width=0.4mm] (-4+2.5*\dx+0.2*\sx, -4.5) -- (-10, -4.5) -- (-10, 4.4) -- (-4+2.5*\dx+0.2*\sx, 4.4) -- (-4+2.5*\dx+0.2*\sx, -4.5); 
		\draw[dashed, color = matcha!150, line width=0.4mm] (-4+2.5*\dx+0.8*\sx, -4.5) -- (12.75, -4.5) -- (12.75, 4.4) -- (-4+2.5*\dx+0.8*\sx, 4.4) -- (-4+2.5*\dx+0.8*\sx, -4.5);

		
		\node at (-4+1.5*\dx+\sx, -5.625+\sy){{\texttt{Input}: Image or Video~ $\Longrightarrow$ ~\texttt{TrIVD}}:  \textbf{\color{cb}Which category should I track} ? or \textbf{\color{matcha!150}Should I detect} ?};
		\draw [mylines, color = orange, line width=0.4mm](-4+1.5*\dx+0.5*\sx, -5.25+\sy) -- (-4+1.5*\dx+0.5*\sx, -5.05+\sy); 
		\draw[dashed, color = orange, line width=0.4mm] (-4+3.5*\dx+\sx, -6+\sy) -- (-4-0.5*\dx+0.5*\sx, -6+\sy) -- (-4-0.5*\dx+0.5*\sx, -5.25+\sy) -- (-4+3.5*\dx+\sx, -5.25+\sy) -- (-4+3.5*\dx+\sx, -6+\sy); 
		\draw [mylines, color = orange, line width=0.4mm](-4-\dx, -5.05+\sy) -- (-4+4*\dx+\sx, -5.05+\sy);

		\node at (-9.75, 4.){\color{cb!90}$\mathbf{t}$};
		\draw [-to, line width=0.6mm, color = cb] (-9.75, 3.65) -- (-9.75, -4.35);

		\draw [decorate,decoration={brace,amplitude=5pt,mirror,raise=6ex},line width=2.pt,color = cb] (-4-1.475*\dx, -3.7) -- (-4-0.525*\dx, -3.7);
		\node at (-4-\dx, -5.15){\texttt{fig/MOT17}~\cite{Milan2016MOT16AB} (tracking)};
		
		\draw [decorate,decoration={brace,amplitude=5pt,mirror,raise=6ex},line width=2.pt,color = gray] (-4-0.475*\dx, -3.7) -- (-4+3.475*\dx+\sx, -3.7);
		\node at (-4+1.5*\dx+0.5*\sx, -5.15){\texttt{fig/VID}~\cite{russakovsky2015vid} (detection - no ground truth tracking annotation available)};
		
		\draw [decorate,decoration={brace,amplitude=5pt,mirror,raise=6ex},line width=2.pt,color = matcha!160] (-4+3.525*\dx+\sx, -3.7) -- (-4+4.475*\dx+\sx, -3.7);
		\node at (-4+4*\dx+\sx, -5.15){\texttt{fig/COCO}~\cite{lin2014coco} (detection)};

		\node at (-4-\dx, 4){\textbf{\color{cb}Track} \textit{``person''}};
		\node at (-4, 4){\textbf{\color{cb}Track} \textit{``airplane''}};
		\node at (-4+\dx, 4){\textbf{\color{cb}Track} \textit{``giant panda''}};
		\node at (-4+2*\dx, 4){\textbf{\color{cb}Track} \textit{``car''}};
		\node at (-4+3*\dx+\sx, 4){\textbf{\color{matcha!150}Detect} in \textit{fig/VIDeo}};
		\node at (-4+4*\dx+\sx, 4){\textbf{\color{matcha!150}Detect} in \textit{image}};

		\draw [myarrows, color = cb](-4-\dx, -5.05+\sy) -- (-4-\dx, -4.45+\sy); 
		\draw [myarrows, color = cb](-4, -5.05+\sy) -- (-4, -4.45+\sy);  
		\draw [myarrows, color = cb](-4+\dx+0.33*\sx, -5.05+\sy) -- (-4+\dx+0.33*\sx, -4.45+\sy); 
		\draw [myarrows, color = cb](-4+2*\dx+0.33*\sx, -5.05+\sy) -- (-4+2*\dx+0.33*\sx, -4.45+\sy); 
		\draw [myarrows, color = matcha!160](-4+3*\dx+\sx, -5.05+\sy) -- (-4+3*\dx+\sx, -4.45+\sy); 
		\draw [myarrows, color = matcha!160](-4+4*\dx+\sx, -5.05+\sy) -- (-4+4*\dx+\sx, -4.45+\sy); 
		
		
		\node at (-4-\dx, -3.35+3*\dy) {\includegraphics[width=3.5cm]{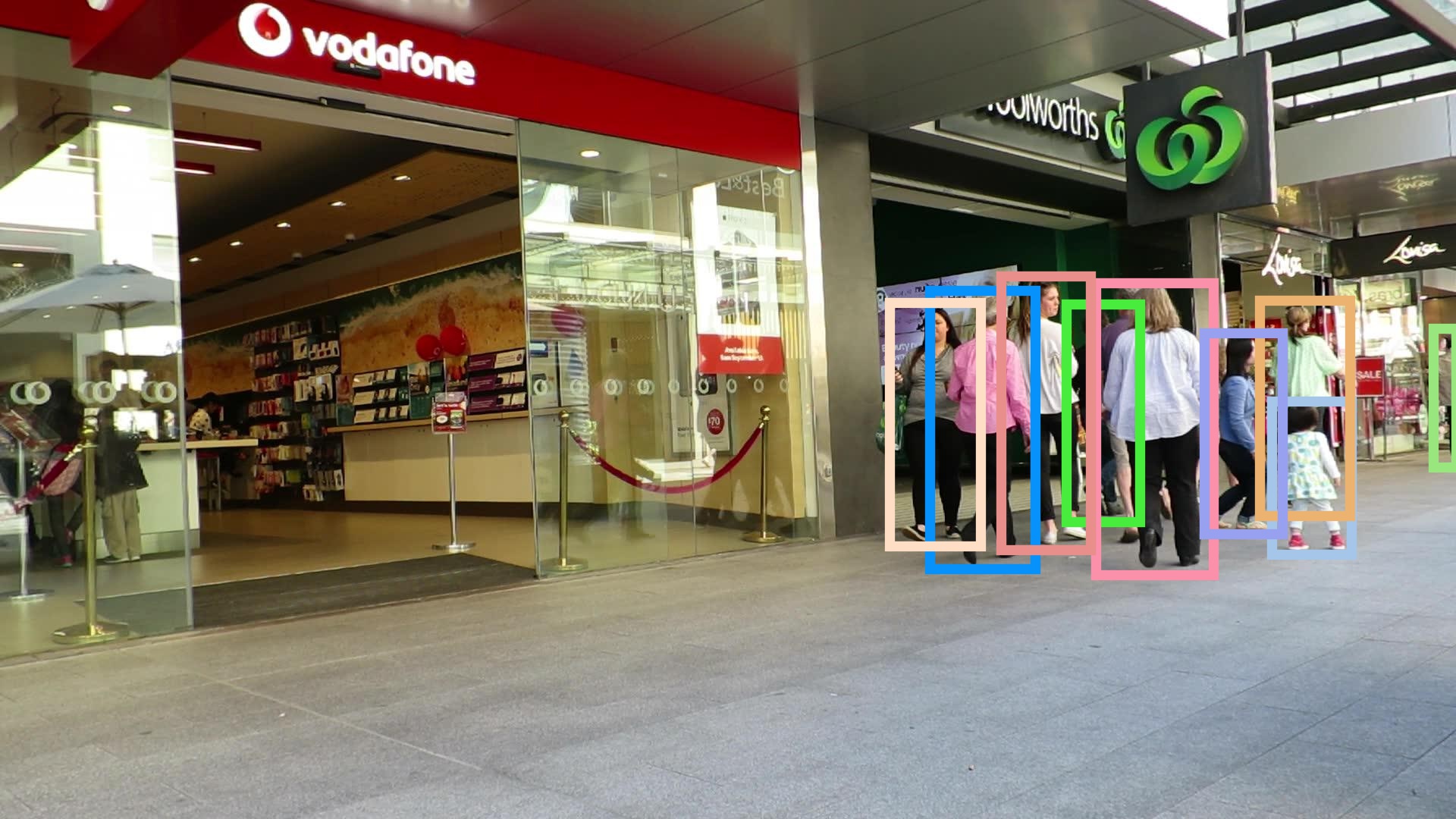}}; 
		\node at (-4-\dx, -3.35+2*\dy) {\includegraphics[width=3.5cm]{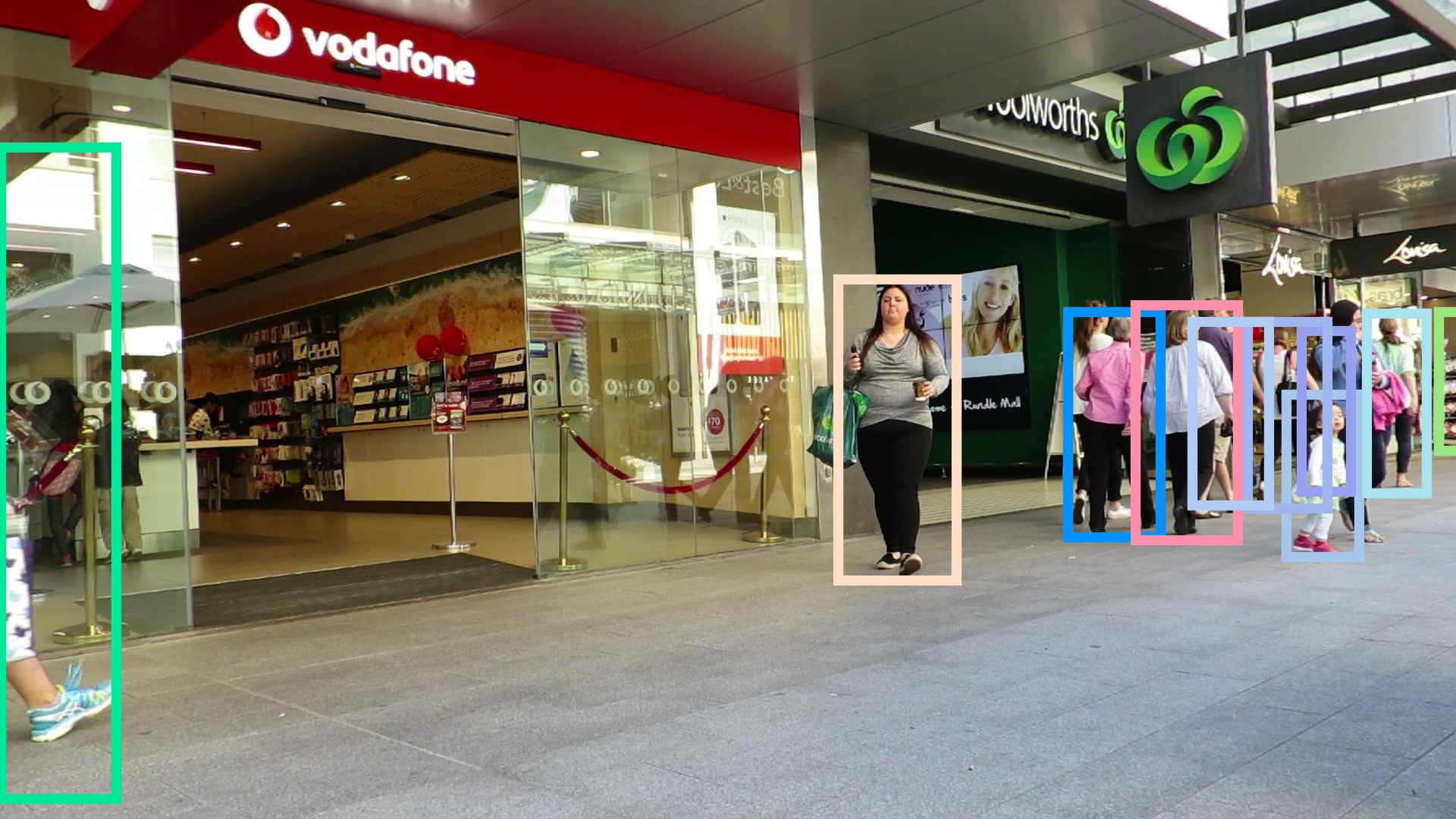}}; 
		\node at (-4-\dx, -3.35+1*\dy) {\includegraphics[width=3.5cm]{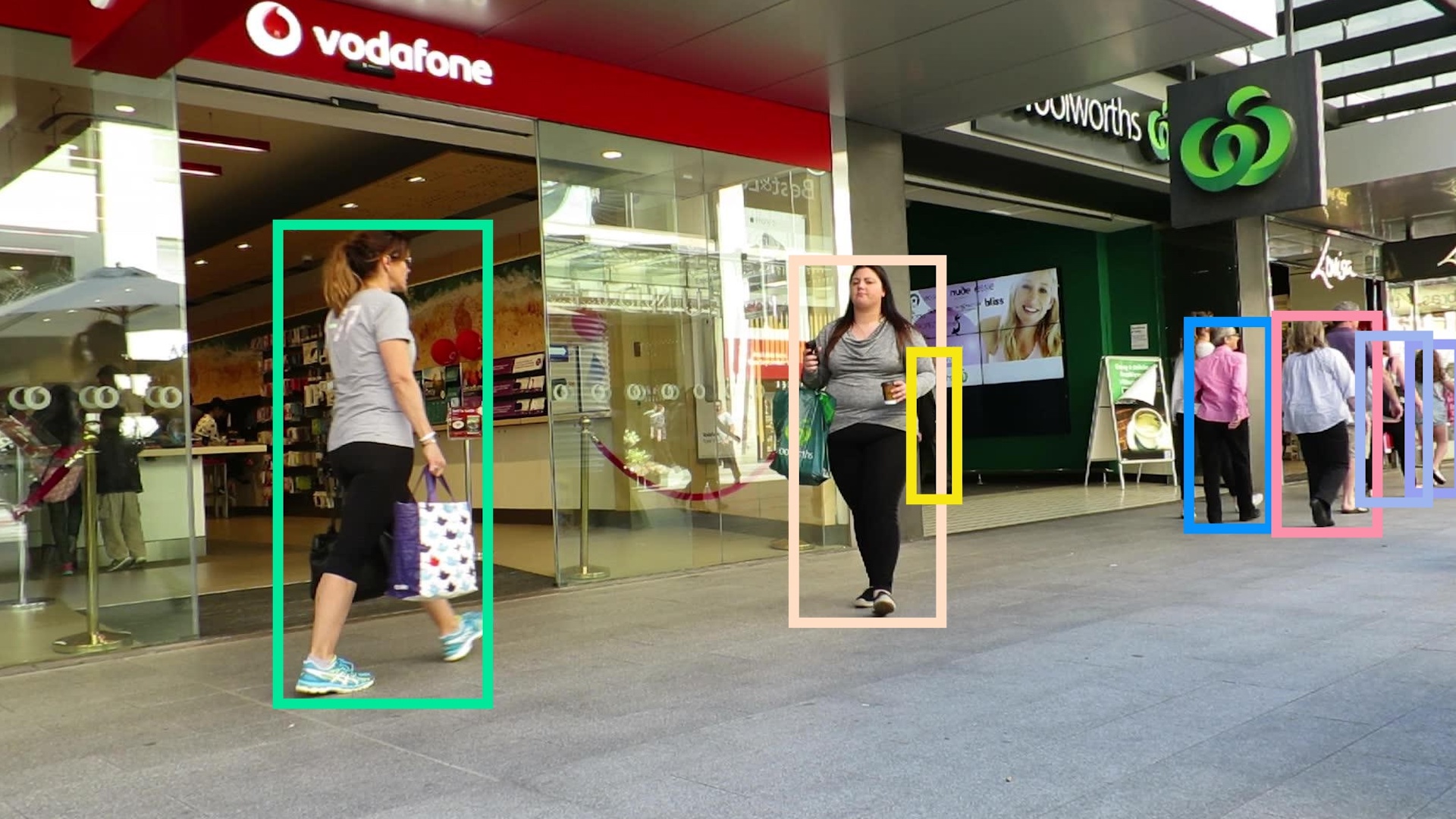}}; 
		\node at (-4-\dx, -3.35) {\includegraphics[width=3.5cm]{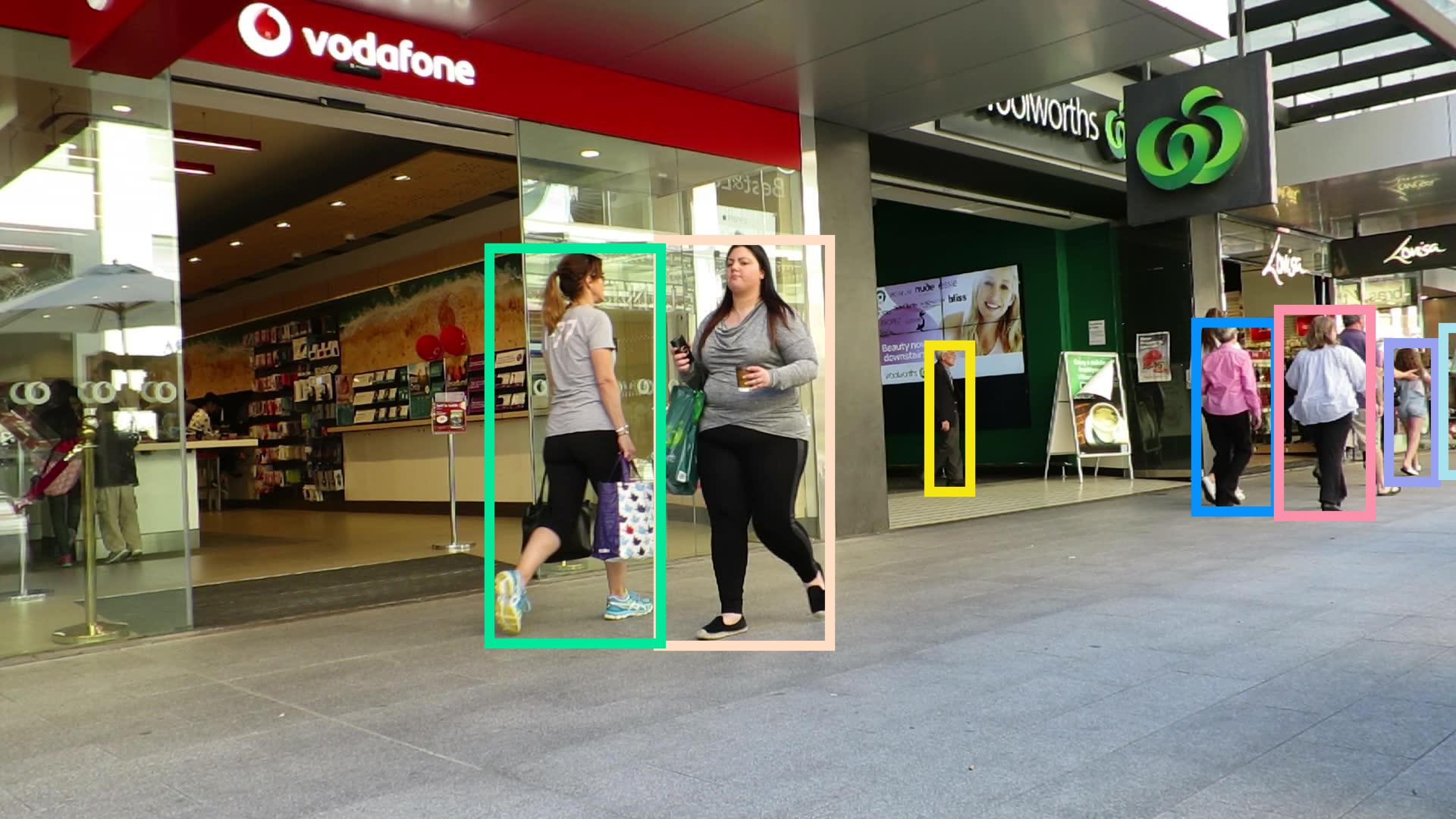}};

		\node at (-4, -3.35+3*\dy) {\includegraphics[width=3.5cm]{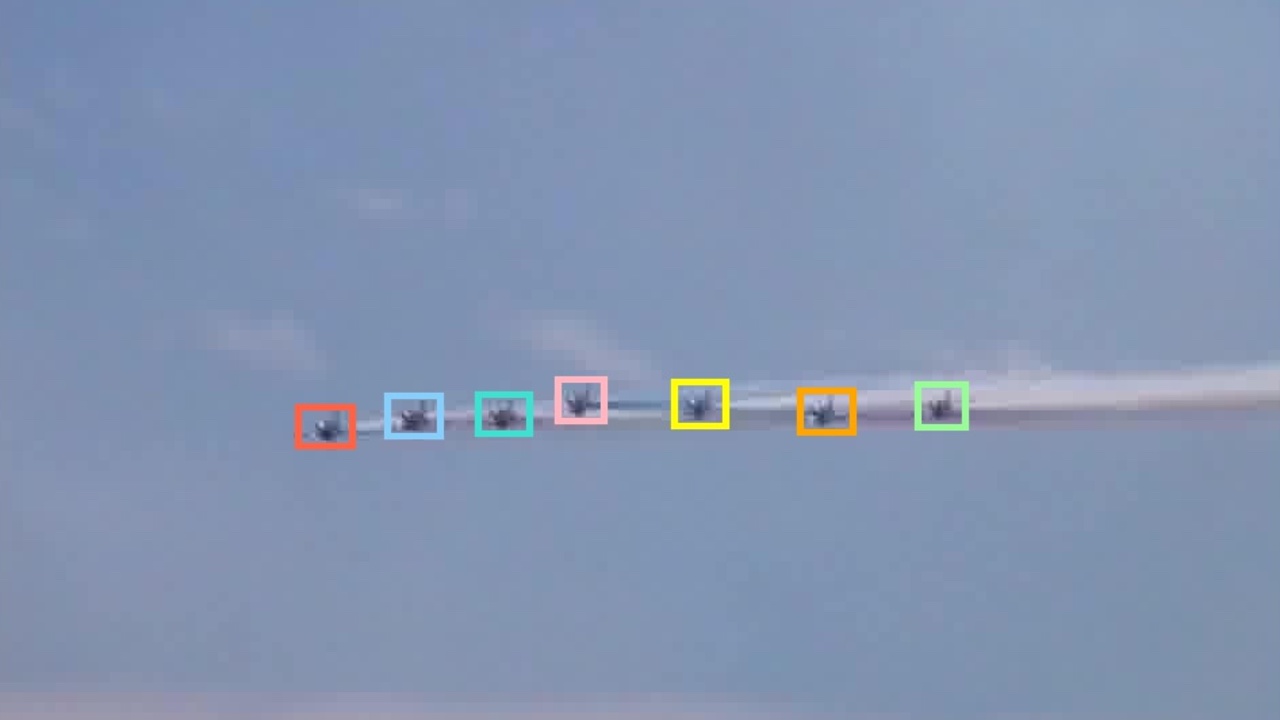}}; 
		\node at (-4, -3.35+2*\dy) {\includegraphics[width=3.5cm]{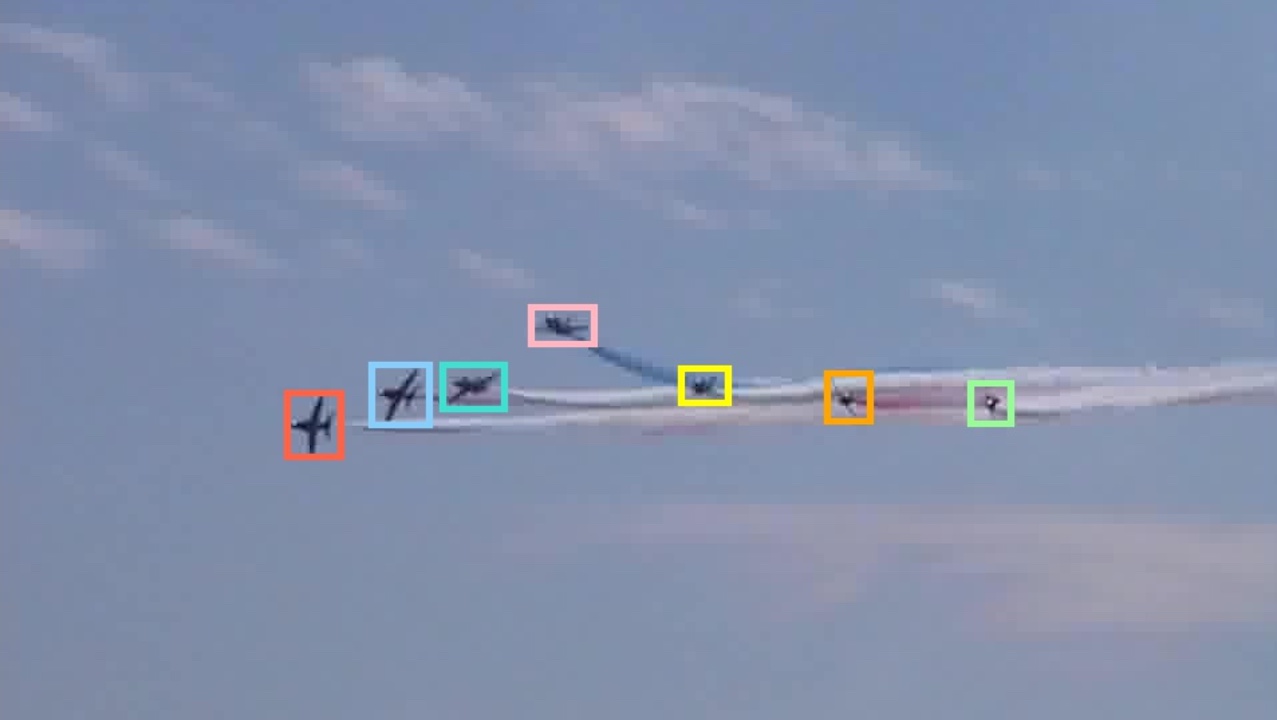}}; 
		\node at (-4, -3.35+1*\dy) {\includegraphics[width=3.5cm]{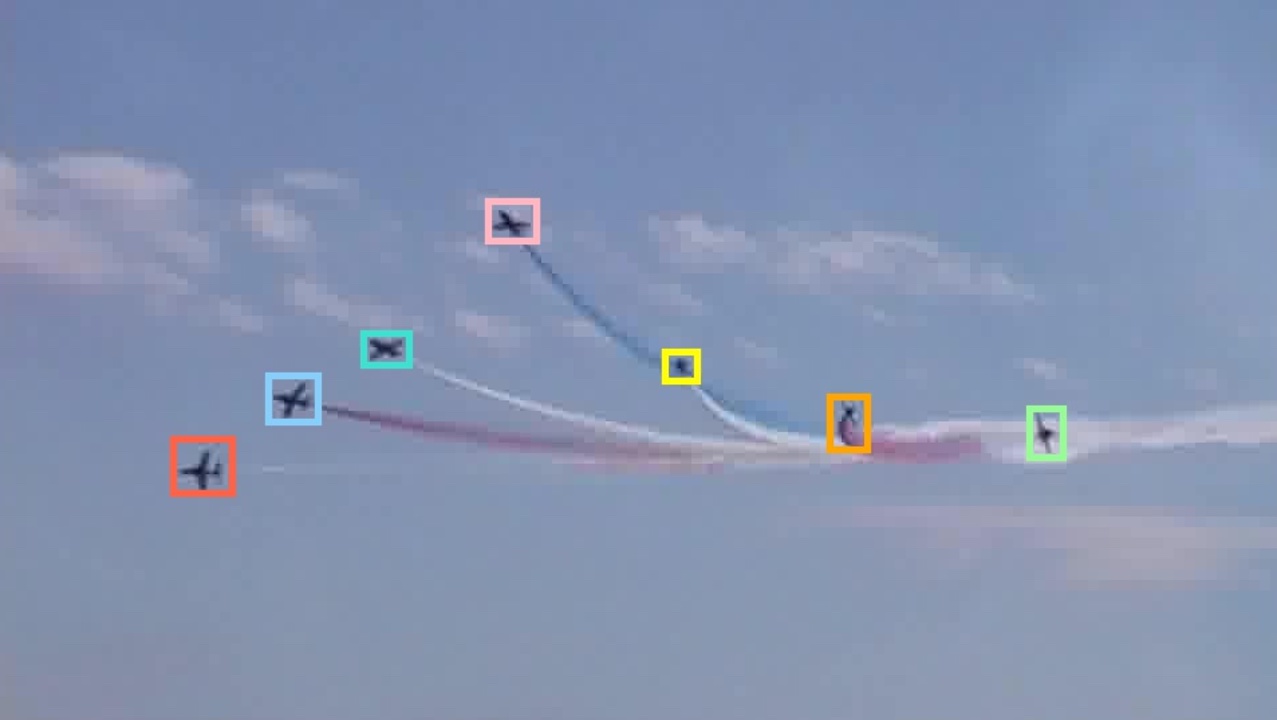}};  
		\node at (-4, -3.35) {\includegraphics[width=3.5cm]{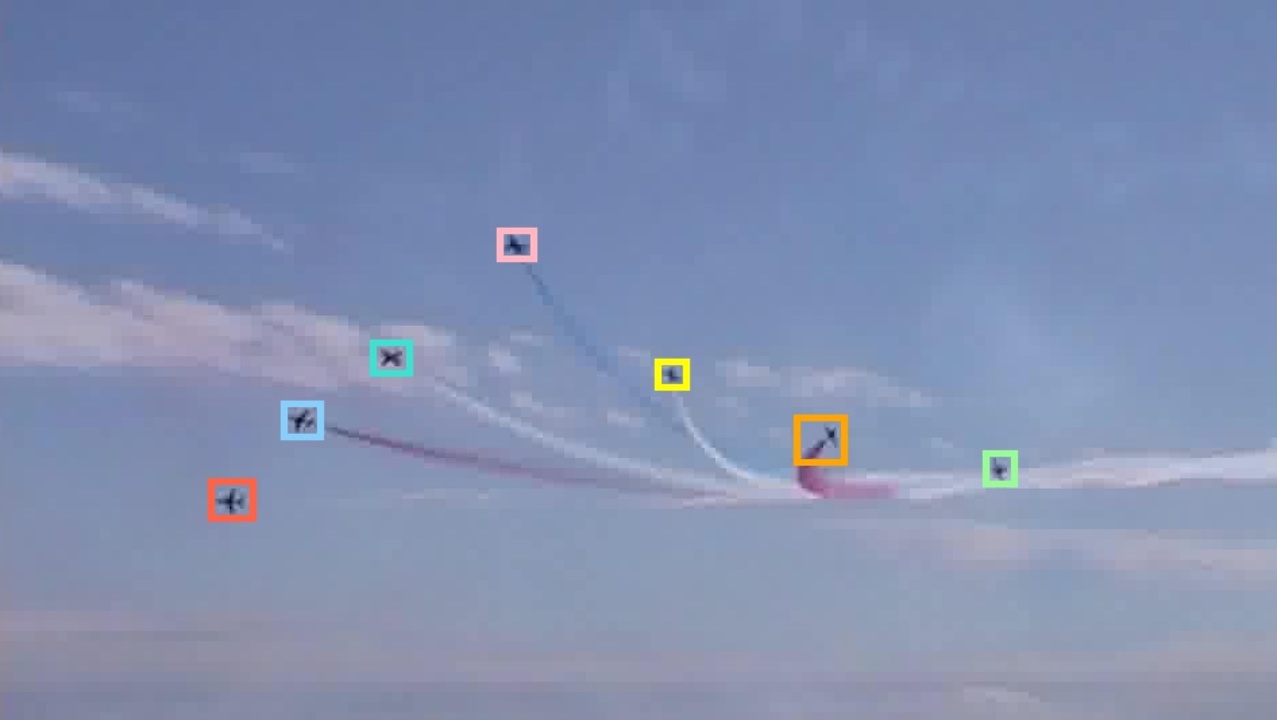}};

		\node at (-4+\dx, -3.35+3*\dy) {\includegraphics[width=3.5cm]{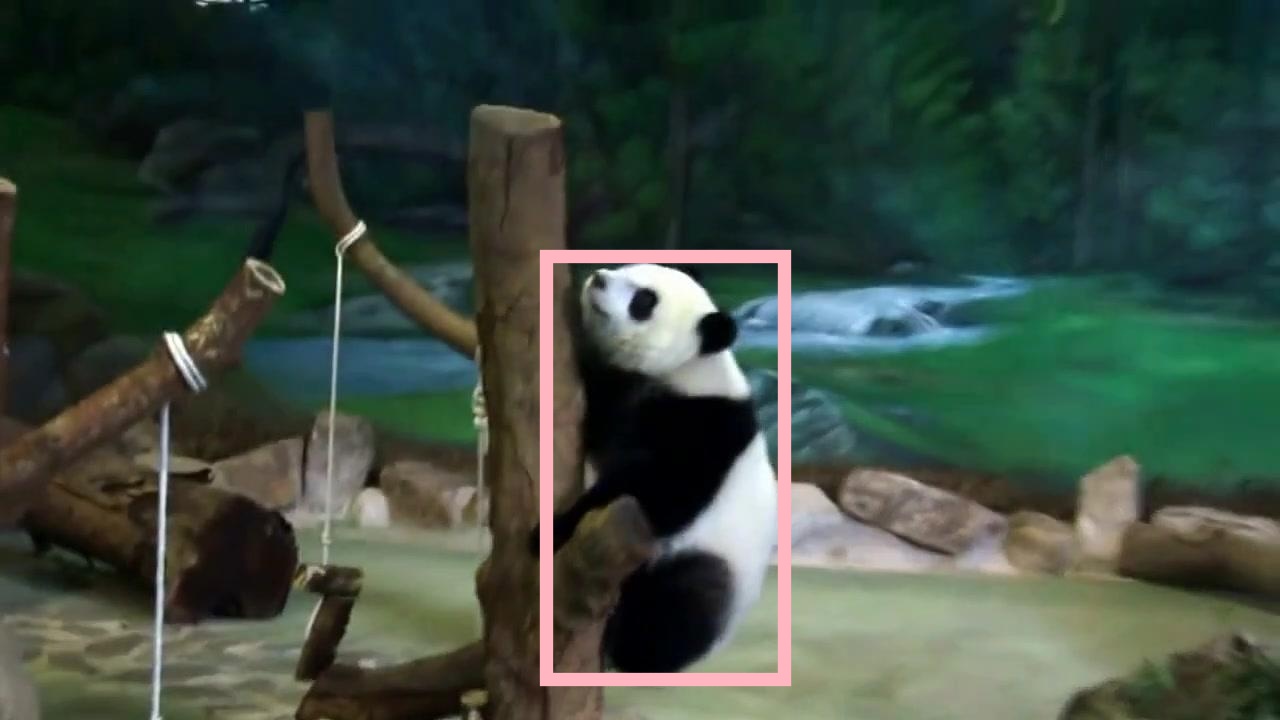}}; 
		\node at (-4+\dx, -3.35+2*\dy) {\includegraphics[width=3.5cm]{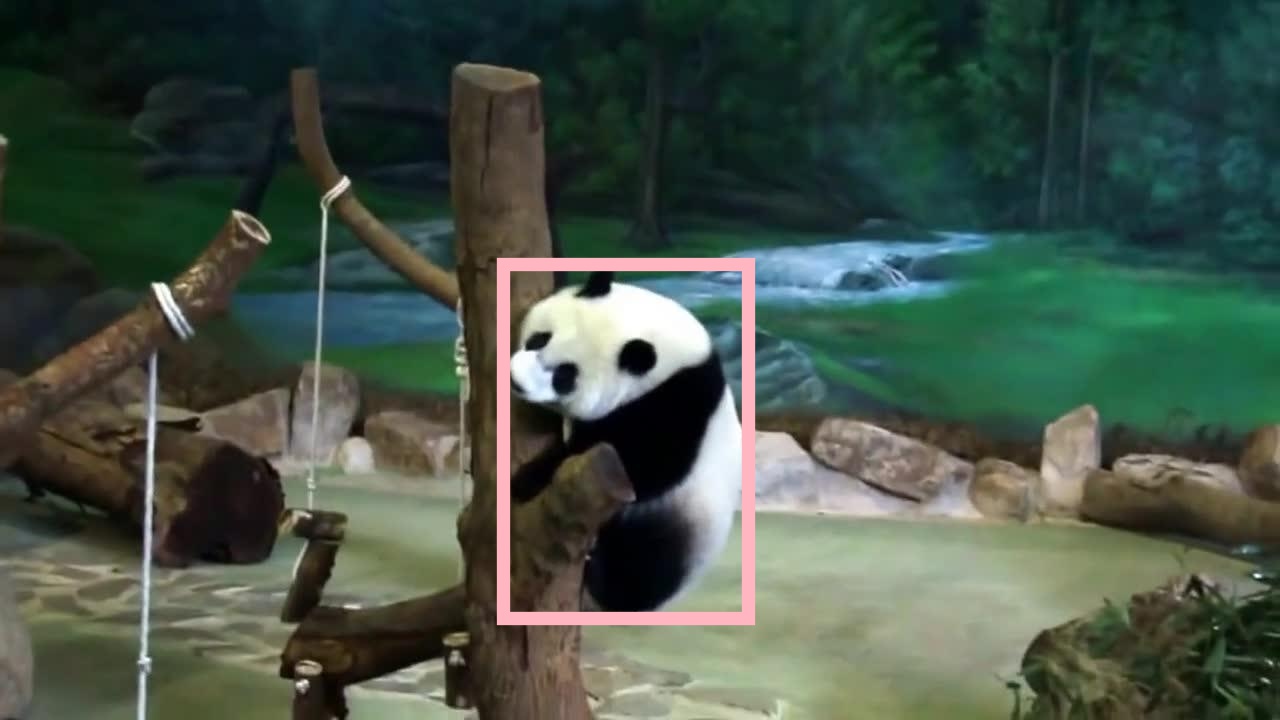}}; 
		\node at (-4+\dx, -3.35+1*\dy) {\includegraphics[width=3.5cm]{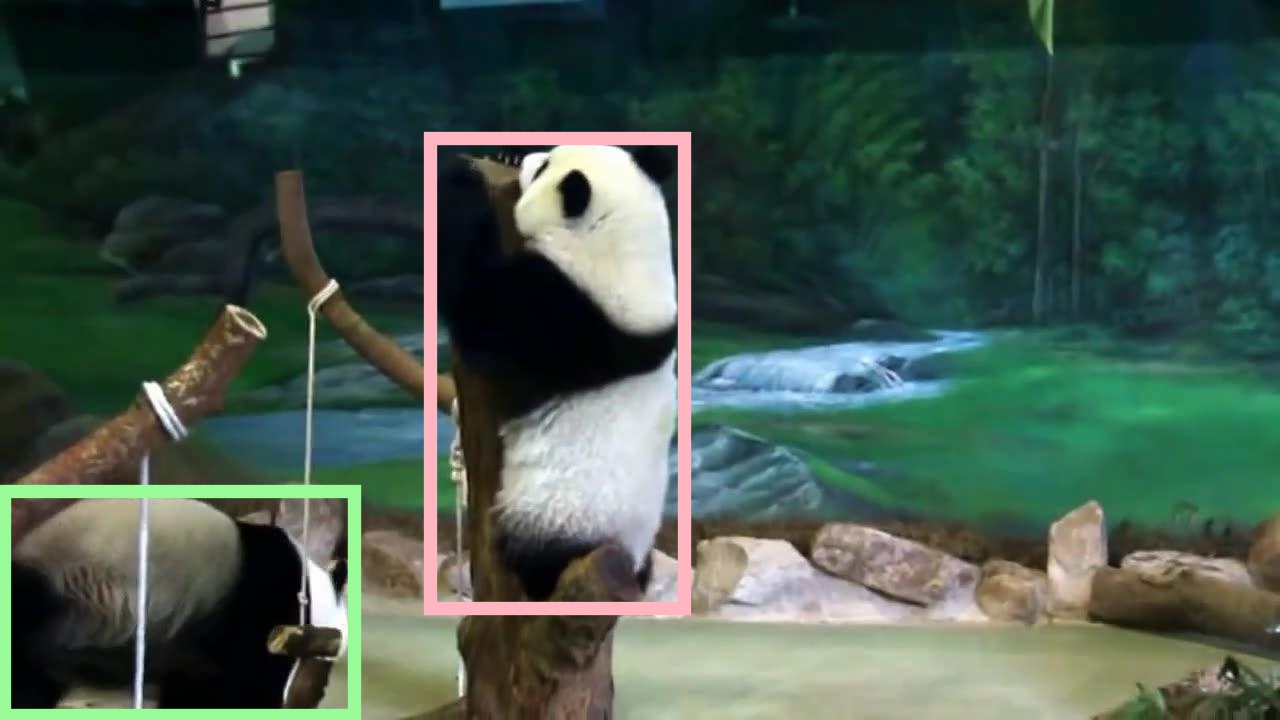}};  
		\node at (-4+\dx, -3.35) {\includegraphics[width=3.5cm]{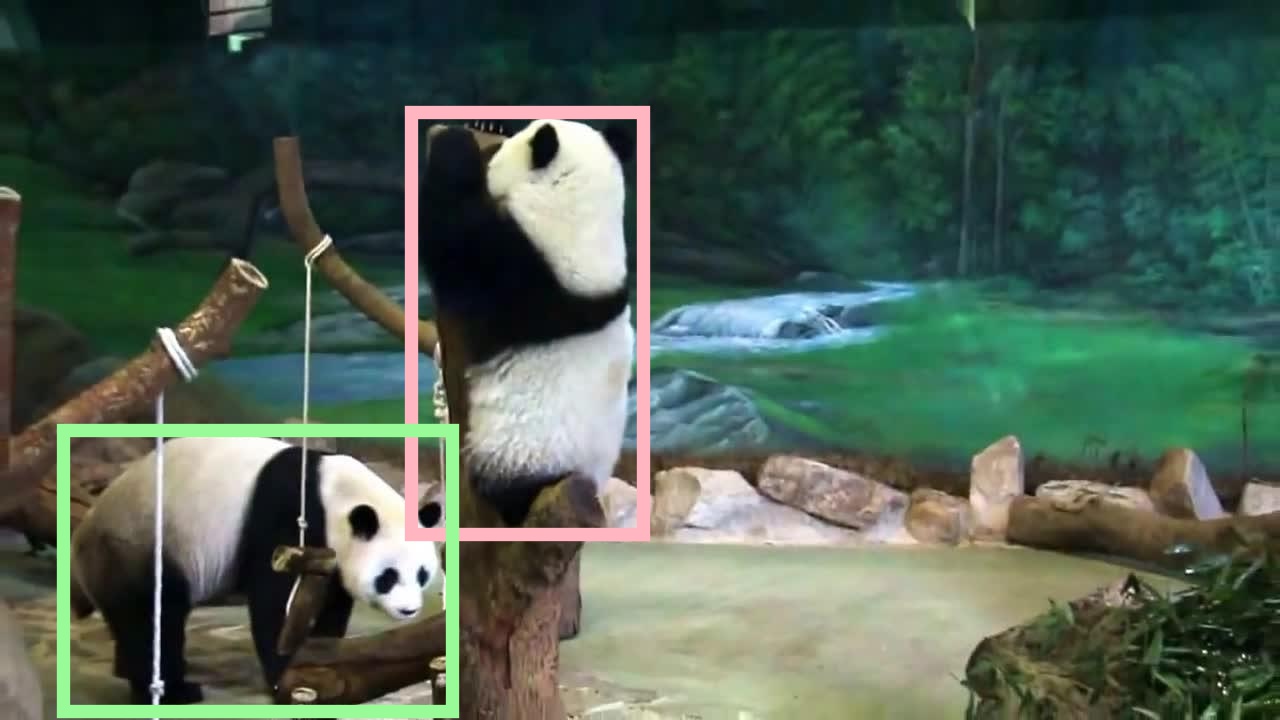}};

		\node at (-4+2*\dx, -3.35+3*\dy) {\includegraphics[width=3.5cm]{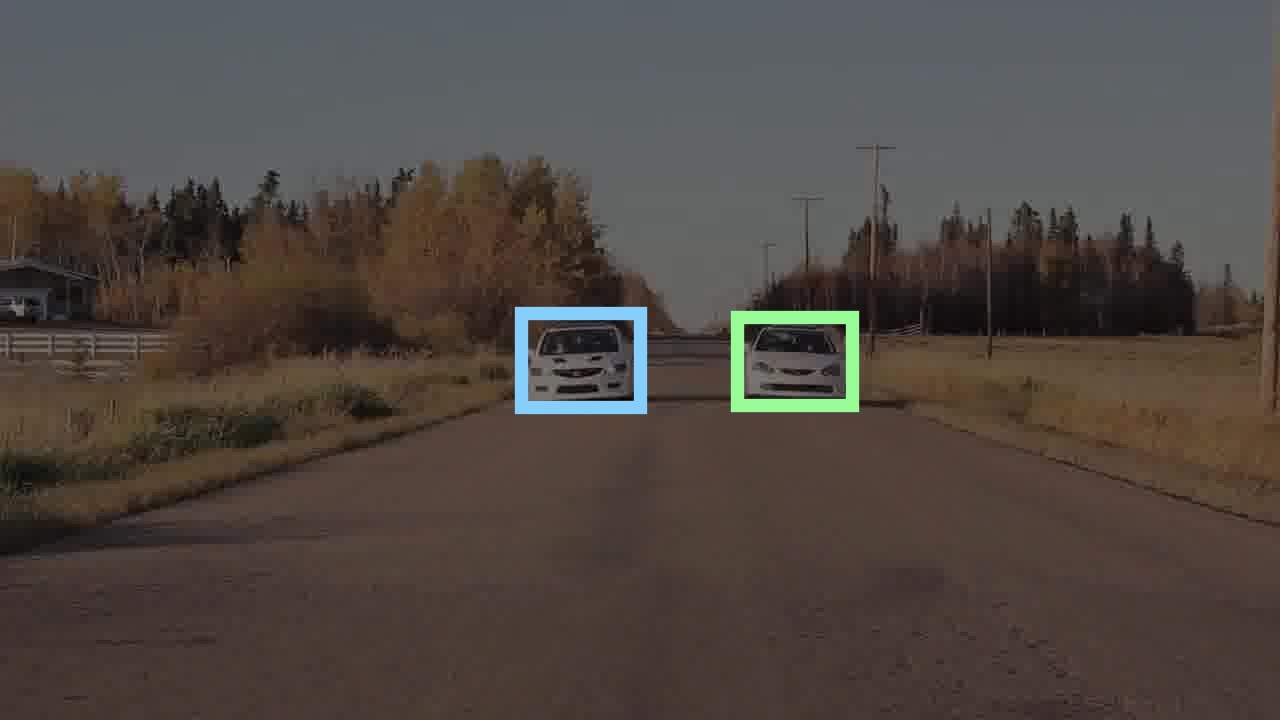}}; 
		\node at (-4+2*\dx, -3.35+2*\dy) {\includegraphics[width=3.5cm]{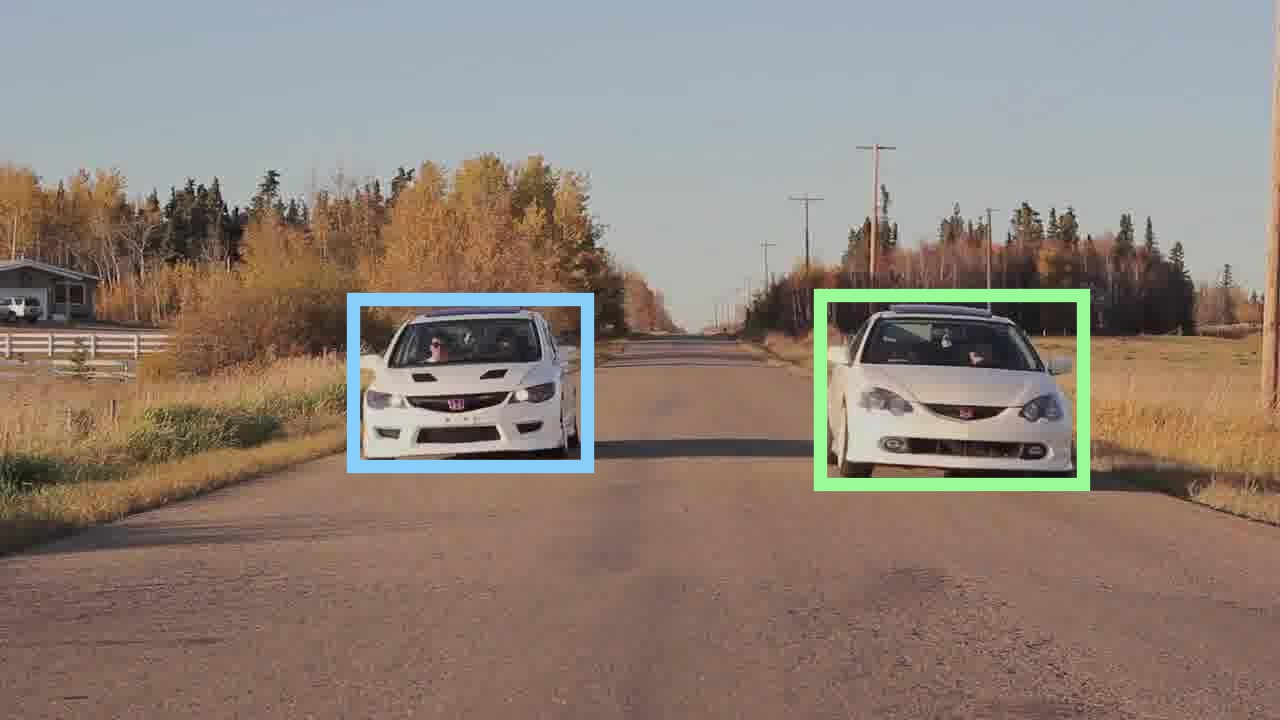}}; 
		\node at (-4+2*\dx, -3.35+1*\dy) {\includegraphics[width=3.5cm]{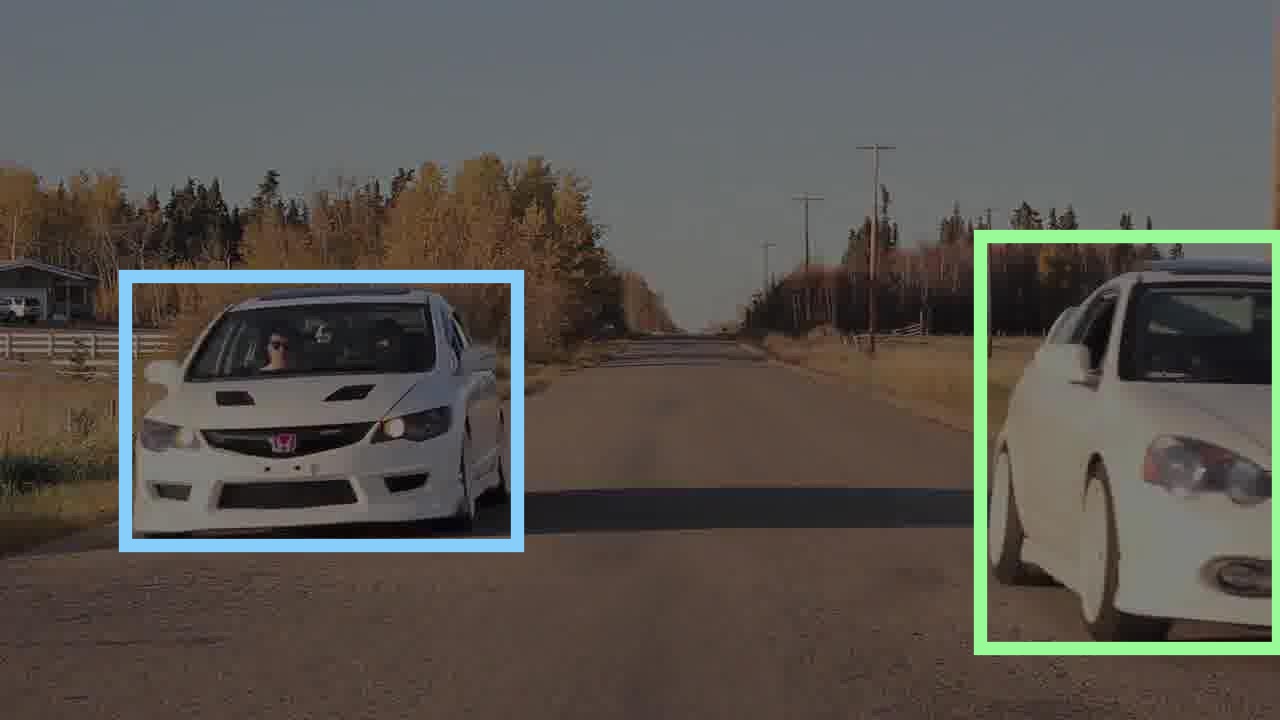}};  
		\node at (-4+2*\dx, -3.35) {\includegraphics[width=3.5cm]{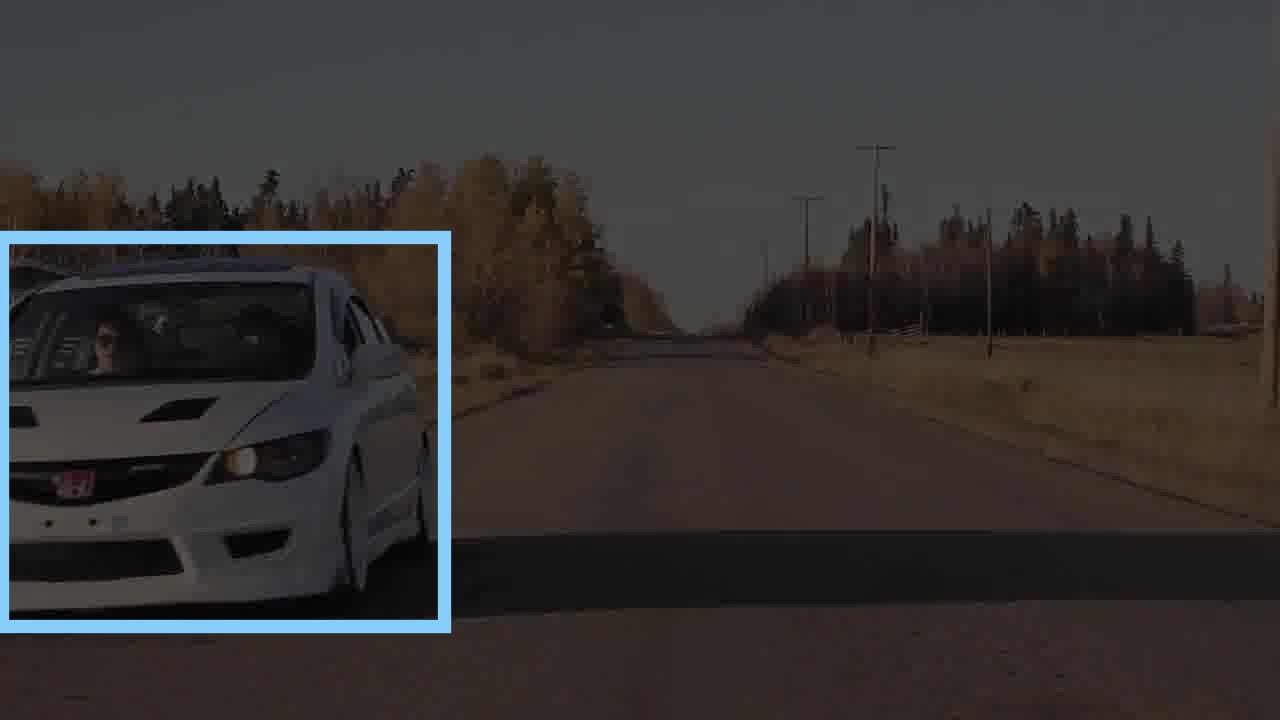}};

		\node at (-4+3*\dx+\sx, -3.35) {\includegraphics[width=3.5cm]{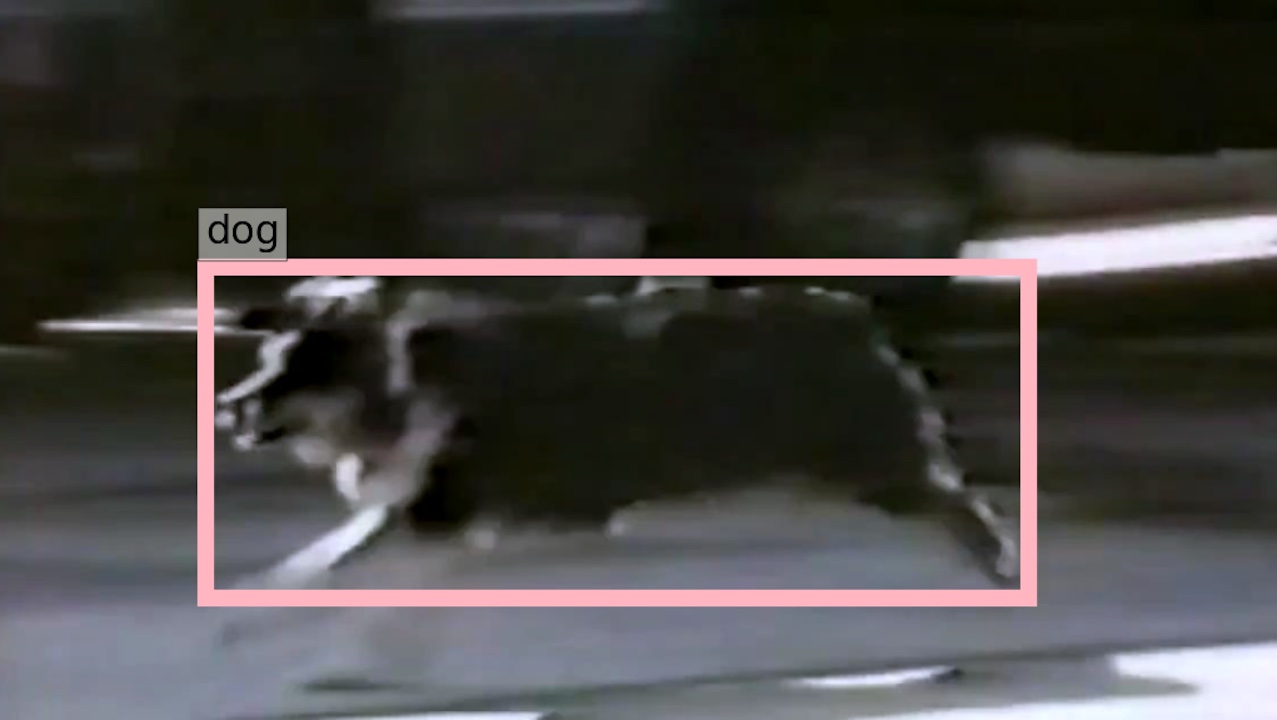}}; 
		\node at (-4+3*\dx+\sx, -3.35+\dy) {\includegraphics[width=3.5cm]{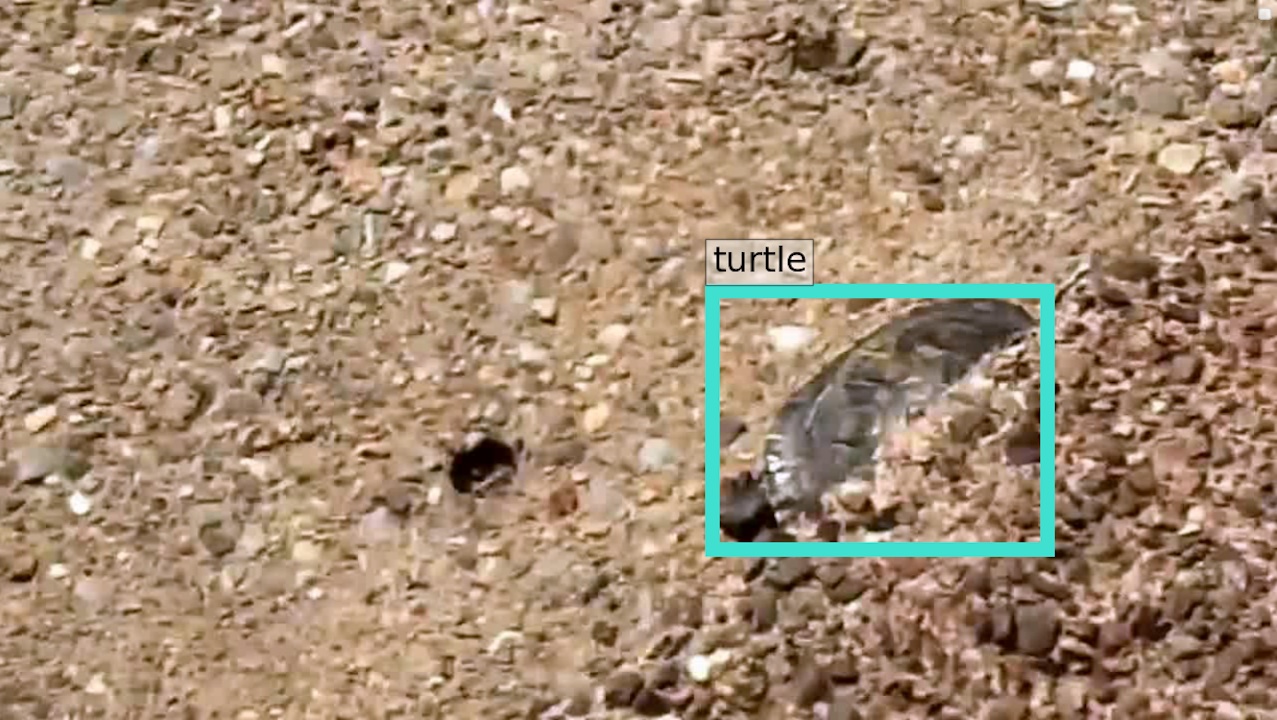}}; 
		\node at (-4+3*\dx+\sx, -3.35+2*\dy) {\includegraphics[width=3.5cm]{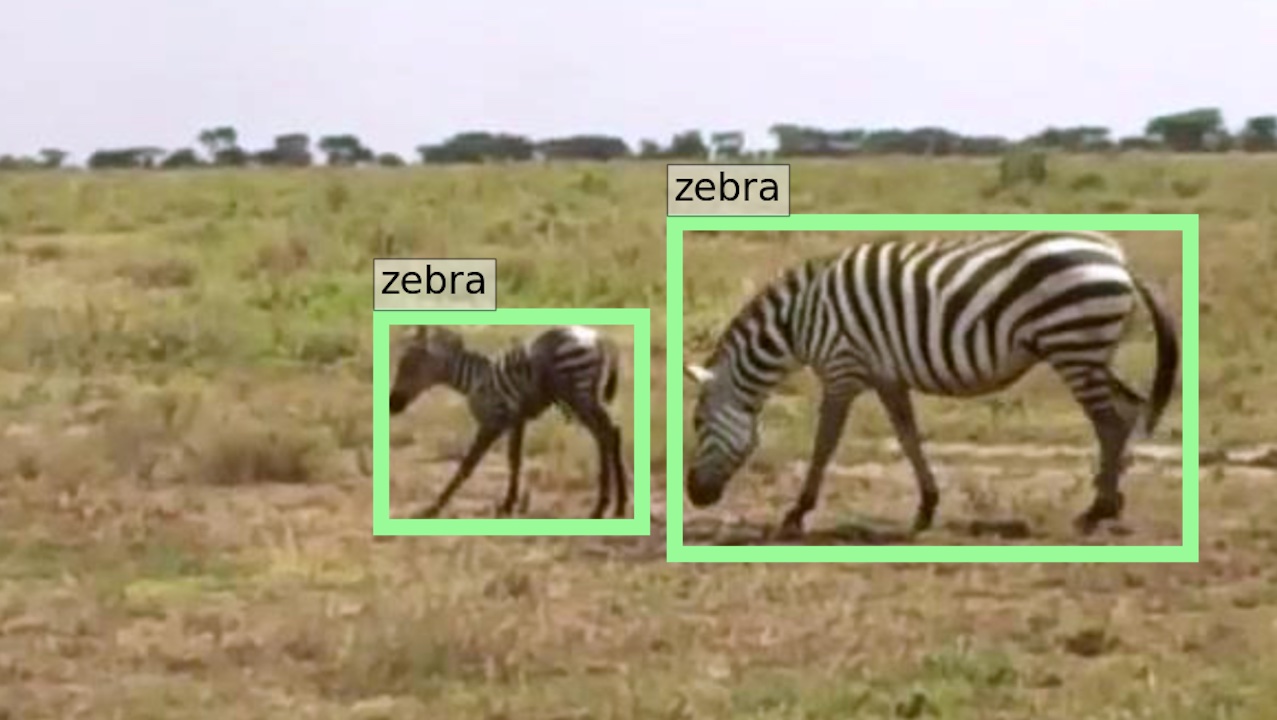}}; 
		\node at (-4+3*\dx+\sx, -3.35+3*\dy) {\includegraphics[width=3.5cm]{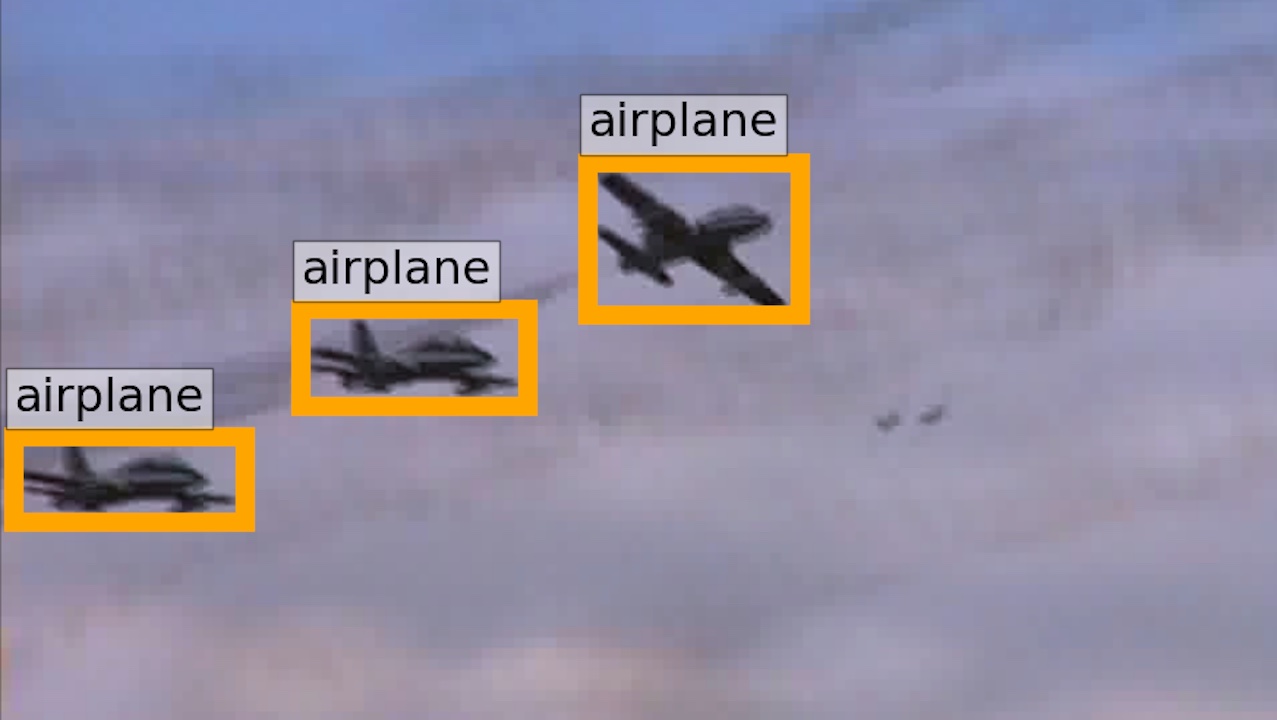}};

		\node at (-4+4*\dx+\sx, -3.35) {\includegraphics[width=3.5cm]{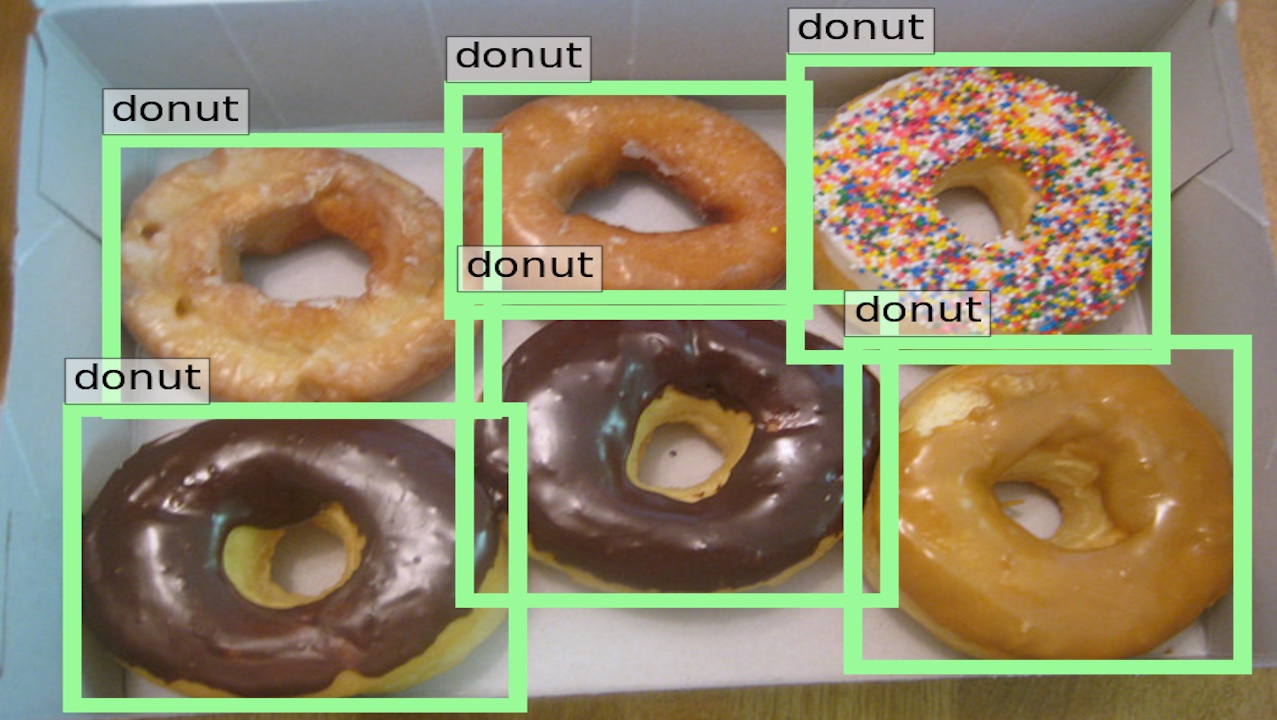}}; 
		\node at (-4+4*\dx+\sx, -3.35+\dy) {\includegraphics[width=3.5cm]{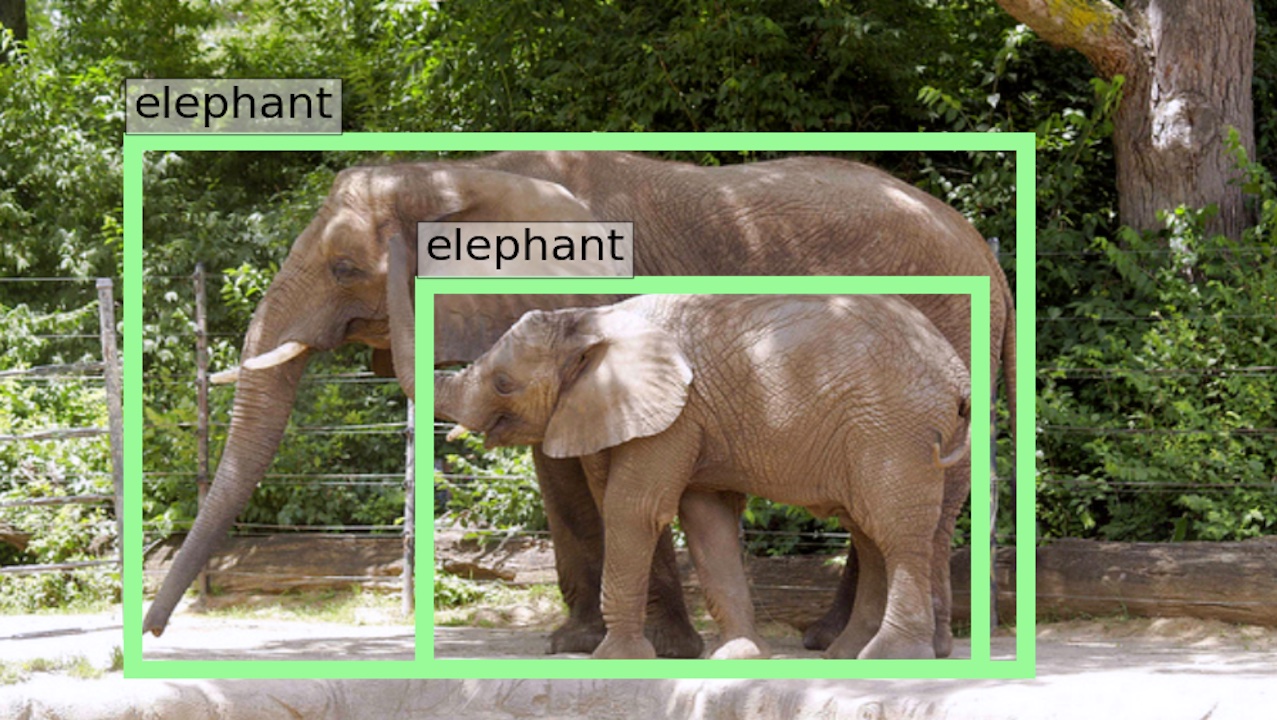}}; 
		\node at (-4+4*\dx+\sx, -3.35+2*\dy) {\includegraphics[width=3.5cm]{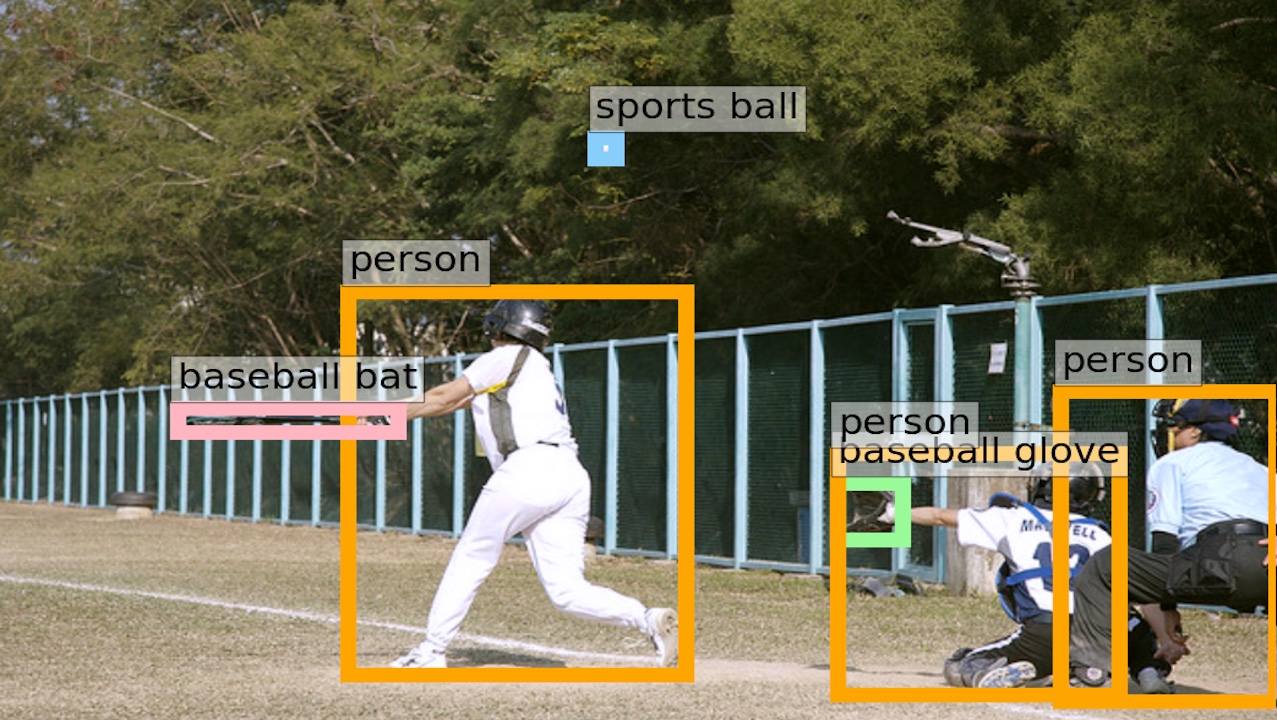}}; 
		\node at (-4+4*\dx+\sx, -3.35+3*\dy) {\includegraphics[width=3.5cm]{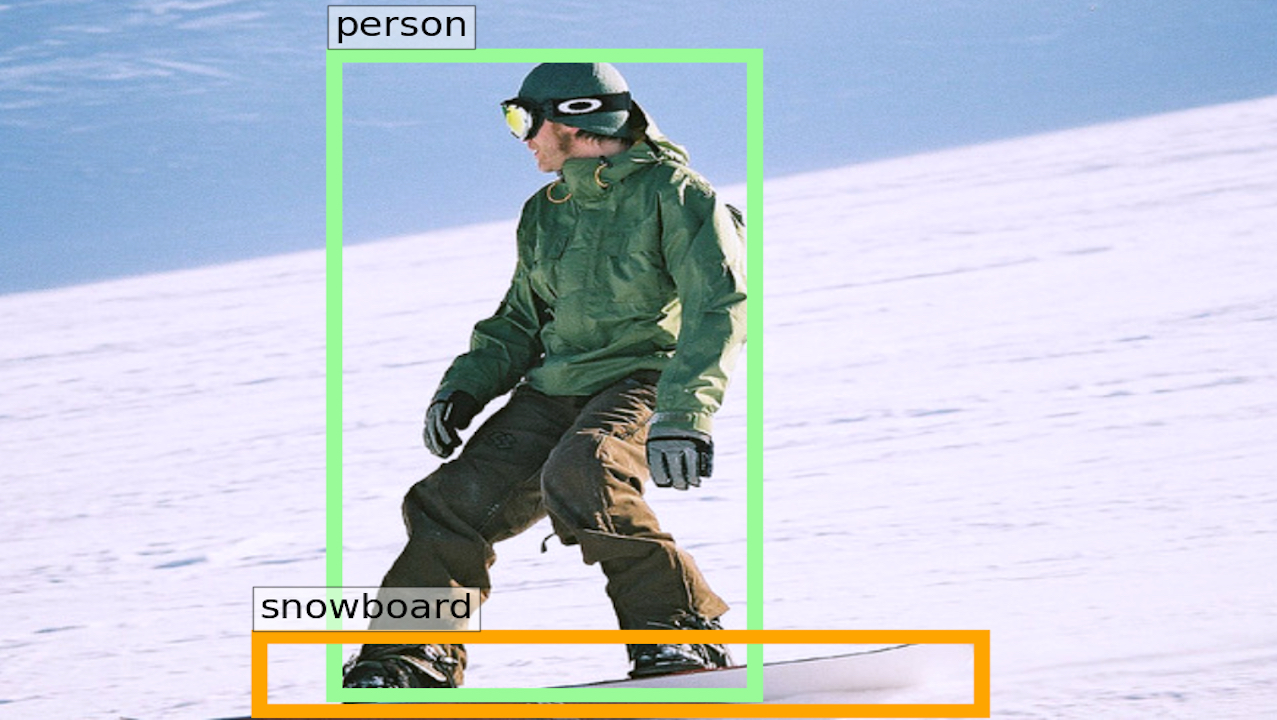}};

	\end{tikzpicture}
	}  
    \caption{\texttt{TrIVD} enables image/video object detection and multi-object tracking within a single model. With the proposed unified framework, we are uniquely able to conduct \textit{zero-shot} multi-object tracking on objects (airplanes, pandas, etc.) that have \textit{not} appeared in tracking datasets. (Different colors refer to object identities in tracking and different object categories in detection figures.)}

	 \label{fig: showcase}
\end{center}

}]

\begin{abstract}
\vspace{-0.3cm}
Objection detection (OD) has been one of the most fundamental tasks in computer vision. Recent developments in deep learning have pushed the performance of image OD to new heights by learning-based, data-driven approaches. On the other hand, video OD remains less explored, mostly due to much more expensive data annotation needs. At the same time, multi-object tracking (MOT) which requires reasoning about track identities and spatio-temporal trajectories, shares similar spirits with video OD. However, most MOT datasets are class-specific (e.g., person-annotated only), which constrains a model's flexibility to perform tracking on other objects. We propose \texttt{TrIVD} (\textbf{Tr}acking and \textbf{I}mage-\textbf{V}ideo \textbf{D}etection), the first framework that unifies image OD, video OD, and MOT within one end-to-end model. To handle the discrepancies and semantic overlaps of category labels across datasets, \texttt{TrIVD} formulates detection/tracking as grounding and reasons about object categories via visual-text alignments. The unified formulation enables cross-dataset, multi-task training, and thus equips \texttt{TrIVD} with the ability to leverage frame-level features, video-level spatio-temporal relations, as well as track identity associations. With such joint training, we can now extend the knowledge from OD data, that comes with much richer object category annotations, to MOT and achieve zero-shot tracking capability. Experiments demonstrate that multi-task co-trained \texttt{TrIVD} outperforms single-task baselines across all image/video OD and MOT tasks. We further set the first baseline on the new task of zero-shot tracking.


\end{abstract}


\section{Introduction}
\label{sec: intro}


Object detection (OD) consists of a localization and classification stage, in which the former determines the location of a potential object and the latter predicts the detected object's category. Traditional detectors address this problem indirectly, by defining surrogate regression and classification problems on a large number of predicted proposals~\cite{ren2015frcnn,cai2021crcnn}, anchors~\cite{lin2020focal}, or window centers~\cite{zhou2019objects,tian2021fcos}. Their performance therefore largely depends on post-processing, e.g., to collapse near-duplicate predictions, design the anchor sets or assign the target boxes to anchors~\cite{zhang2020bridging}. DETR-based methods~\cite{carion2020detr,zhu2020deformable,he2021end,minderer2022sov}, as fully end-to-end object detectors, eliminate the need for hand-crafted components via the relation modeling capability of vision transformers (ViTs)~\cite{dosovitskiy2020vit}. Coupled with language encoders, 
recent open-vocabulary detection models are further able to leverage information from the large amounts of image/object-text data, to boost  performance and further achieve zero-shot detection~\cite{kamath2021mdetr,zhou2021detic,gu2022vild,li2021glip,Zhang2022GLIPv2UL,minderer2022sov,zang2022ovdetr}. Resorting to region-text alignment, we achieve more flexibility in handling cross-dataset discrepancies~(Sec.~\ref{sec: vl}) and open up possibilities for open-vocabulary zero-shot tracking~(Sec.~\ref{sec: open-vocab}).


However, the above developments mainly focus on \textit{image} OD, leaving \textit{video} OD less-explored, largely because video OD models usually have many bespoke hand-crafted components, e.g., optical flow~\cite{zhu2017dff,zhu2017fgfa,zhu2018thp} which requires 
additional flow data. Another major challenge lies in applying advanced modern architectures like ViTs on the high-resolution, space-time videos due to the high computational cost from self-attention's quadratic complexity~\cite{vaswani2017attn}. TransVOD~\cite{he2021end} uses two ViTs, one extracts frame-level features, the other leverages temporal relations. This strategy still faces quadratically increasing self-attention computations w.r.t. input videos' temporal lengths. By uniformly formulating images/videos as frame-level inputs followed by sequential attentions~(Sec.~\ref{sec: dynhead}), we reduce the computation time while achieving better performance~(Tab.~\ref{tab: omnivore}).

Meanwhile, multi-object tracking (MOT) models the tracking identities and the spatio-temporal trajectories~\cite{meinhardt2021trackformer}, which shares similar goals with OD in general on locating potential objects, and with video OD in particular on reasoning about spatio-temporal relations between adjacent frames. Recent advances in MOT approaches mainly pursue tracking by detection~\cite{Kim2015MultipleHT,LealTaix2016LearningBT,Henschel2017ImprovementsTF,Chen2018RealTimeMP,Keuper2020MotionS,LealTaix2011EverybodyNS}, by regression~\cite{Bergmann2019TrackingWB,Braso2020LearningAN,Zhou2020TrackingOA,Liu2020GSMGS,Dendorfer2021MOTChallengeAB}, or by attention~\cite{meinhardt2021trackformer,Zhang2022ByteTrackMT,Yan_2021_ICCV,Zeng2022MOTREM}. Built upon the recent advances of ViTs~\cite{dosovitskiy2020vit,vaswani2017attn}, tracking-by-attention~\cite{zhu2020deformable,meinhardt2021trackformer} associates objects across frames via the self-attention intrinsically produced by ViTs, and naturally relates tracking with frame-level detection. We follow the tracking-by-attention mechanism. By inheriting the proposed objects from previous frames, we achieve detection and data association in tracking simultaneously~(Sec.~\ref{sec: det_track}).


We present a unified framework, \texttt{TrIVD} (\textbf{Tr}acking and \textbf{I}mage-\textbf{V}ideo \textbf{D}etection), which incorporates the three (image/video OD, MOT) tasks in one end-to-end model. \texttt{TrIVD} could be trained on image/video OD and MOT datasets individually, or co-trained in a cross-dataset, multi-task fashion. We highlight our contributions as follows:

\paragraph{Bridging the gap between image/video OD and MOT.}
Existing OD models are specifically designed for either image or video OD task with insufficient \textit{flexibility} in handling inputs containing both images and videos. \texttt{TrIVD} formulates image and video inputs uniformly, with an integrated \textit{temporal-aware attention} module to efficiently leverage the spatio-temporal relations for video inputs~(Sec.~\ref{sec: dynhead}, Tab.~\ref{tab: omnivore}).

We further formulate MOT in a tracking-by-attention fashion~\cite{meinhardt2021trackformer}, where detection and tracking data association are performed jointly via self-attention without additional track matching procedures~(Sec.~\ref{sec: det_track}). This enables the multi-dataset, multi-task training of \texttt{TrIVD}, and equips \texttt{TrIVD} with zero-shot tracking ability to track objects that have \textit{not} been seen in MOT training data~(Sec.~\ref{sec: zero-shot}).

\paragraph{Setting the first zero-shot tracking baseline.} To the best of our knowledge, \texttt{TrIVD} is the first work achieving zero-shot tracking. Besides qualitative visualizations~(Fig.~\ref{fig: showcase}, Fig.~\ref{fig: mot}), we further conduct quantitative evaluations and set the first baseline on the new zero-shot tracking task (Tab.~\ref{tab: tao}).

\paragraph{Unifying multi-task datasets with region-text alignment.} Class categories vary across different OD/MOT datasets, yet \textit{semantic} overlaps may still exist. We re-formulate the category prediction in OD/MOT via phrase grounding~\cite{li2021glip} such that \texttt{TrIVD} is given both image/video inputs \textit{and} a text prompt containing all the candidate categories to be detected/tracked~(Sec.~\ref{sec: vl}). By aligning visuals with their semantic meanings, we intrinsically resolve the class label discrepancies and semantic overlaps across datasets. 

Trained in a multi-task, multi-dataset fashion, \texttt{TrIVD} not only outperforms single-task baselines across all image OD, video OD and MOT tasks~(Sec.~\ref{sec: compare}), but also uniquely achieves \textit{zero-shot} tracking performance, and thus is able to track objects without the need of being trained with their ground truth tracking identity annotations~(Sec.~\ref{sec: zero-shot}).

\section{Related Work}
\label{sec: related_work}

\paragraph{Image Object Detection (Image OD)}

detects objects with their categories~\cite{huang2021survey}. With the advent of convolutional neural networks (CNNs), current leading object detectors are built upon CNNs~\cite{Krizhevsky2012cnn,Simonyan2015VeryDC,Szegedy2015dcnn,He2016DeepRL,Cao2019gcnet} and can be generally classified into two main categories: anchor-based detectors (e.g., R-CNN~\cite{Girshick2014rcnn}, Fast(er) R-CNN~\cite{Girshick2015frcnn,ren2015frcnn}, Cascade R-CNN~\cite{cai2021crcnn}) and anchor-free detectors (e.g., CornerNet~\cite{Law2019cornernet}, ExtremeNet~\cite{zhou2019bottomup}). The former can be divided into two-stage and one-stage methods, the latter falls into the class of keypoint-based and center-based methods~\cite{zhang2020bridging}. 

Transformers~\cite{carion2020detr,dosovitskiy2020vit,meinhardt2021trackformer,Sun2021sprcnn,wang2020end,zhu2020deformable} have received great attention recently. DETR-based methods~\cite{carion2020detr,zhu2020deformable} build end-to-end object detection models based on Transformers, and largely simplify the detection pipeline~\cite{ren2015frcnn}. Coupling with language encoders and contrastive learning~\cite{radford2021clip}, a new stream of open-vocabulary detection works take advantage of the large amounts of image/object-text grounding data, further boost the model's performance and achieve zero-shot capabilities~\cite{kamath2021mdetr,zhou2021detic,gu2022vild,li2021glip,Zhang2022GLIPv2UL,minderer2022sov,zang2022ovdetr}. \texttt{TrIVD}, an end-to-end, unified model for image/video OD and MOT~(Fig.~\ref{fig: showcase}), builds upon  deformable-DETR~\cite{zhu2020deformable} and resorts to region-text alignment for a unified classifier.


\paragraph{Video Object Detection (Video OD)} detects per-frame objects as image OD, usually requires cross-frame linking for occluded or blurred objects. A common solution~\cite{chen2018stlattice,chen2020mega,Guo2019ProgressiveSL,han2020favid,han2020hvrnet,He2020TemporalCE,jiang2020learning,Lin2020DualSF,Sun2021MAMBAMA,Yao2020VideoOD} is using feature aggregation to enhance per-frame features by aggregating the features of nearby frames with flow-based warping~\cite{zhu2017dff,zhu2017fgfa,zhu2018thp,Dosovitskiy2015flownet}. Another line of 
methods utilize self-attention~\cite{vaswani2017attn} and non-local information~\cite{wang2018nonlocal} to capture the long-range dependencies of temporal contexts~\cite{wu2019selsa,Bertasius2018ObjectDI,He2020TemporalCE,Deng2019RelationDN,jiang2019lwdn,deng2019ogemn,chen2020mega}. However, most pipelines for video OD are sophisticated and include multiple post-processing steps with hand-crafted components~\cite{Han2016SeqNMSFV,kang2018tubelet,Belhassen2019ImprovingVO,sabater2020repp}. TransVOD~\cite{he2021end} applies vision transformers (ViTs) to build an end-to-end video OD model, and handles the spatio-temporal relations by using a ViT to extract frame-level features, and an additional ViT for temporal aggregation, which results in a quadratic increase in self-attention computation along the temporal axis. In contrast, \texttt{TrIVD} uniformly formulates image and video inputs with its temporal-aware sequential attentions, which efficiently fuses features across video frames~(Sec.~\ref{sec: dynhead}, Tab.~\ref{tab: omnivore}).

\paragraph{Multi-object Tracking (MOT)} models spatio-temporal trajectories of tracking identities~\cite{meinhardt2021trackformer}. Recent works generally focus on three aspects, tracking by detection, by regression, or by attention. Tracking-by-detection detects frame-wise objects and then associats the object identities across adjacent frames~\cite{Kim2015MultipleHT,LealTaix2016LearningBT,Henschel2017ImprovementsTF,Chen2018RealTimeMP,Keuper2020MotionS,LealTaix2011EverybodyNS}. Tracking-by-regression applies a continuous regression following the positions of detected objects between frames~\cite{Bergmann2019TrackingWB,Braso2020LearningAN,Zhou2020TrackingOA,Liu2020GSMGS,Dendorfer2021MOTChallengeAB}. Tracking-by-attention associates objects via self-attention~\cite{dosovitskiy2020vit,meinhardt2021trackformer,Zhang2022ByteTrackMT,Yan_2021_ICCV,Zeng2022MOTREM,Chu2021TransMOTSG,Zhou2022GTR}, and naturally relates frame-level tracking and detection. We follow tracking-by-attention approaches, and integrate detection and tracking in our unified framework~(Sec.~\ref{sec: det_track}). Co-trained on image/video OD and MOT datasets, \texttt{TrIVD} not only can perform all the three tasks, but also is able to track \textit{novel} object categories without the need for supervised training on their tracking annotations~(Sec.~\ref{sec: zero-shot}, Fig.~\ref{fig: mot}).

\paragraph{Multi-dataset, Multi-modal and Multi-task Learning} Multi-modal learning architectures allow training separate encoders for different input modalities, such as image-text~\cite{Castrejon2016multimodal,Gong2014ImprovingIE,Karpathy2017vsalign,lu2020multitask,Miech2020EndtoEndLO}, video-audio~\cite{Arandjelovi2017LookLA,Arandjelovi2018ObjectsTS,Morgado2021RobustAI,Morgado2021AudioVisualID,Owens2018AudioVisualSA,Patrick2020MultimodalSF}, video-optical flow~\cite{Simonyan2014TwoStreamCN}, etc. Most multi-modal models assume the input modalities are in correspondence and available simultaneously, while \texttt{TrIVD} operates on multi-modal inputs but does \textit{not} require access to all modalities. Multi-task learning~\cite{Caruana2004MultitaskL} operates on the same modality but output predictions for multiple tasks~\cite{Eigen2015PredictingDS,Ghiasi2021MultiTaskSF,Kokkinos2017UberNetTA,Maninis2019AttentiveSO,Misra2016CrossStitchNF,Zhang2014FacialLD}, while \texttt{TrIVD} is able to handle both image and video inputs, and conduct all OD and MOT tasks~(Fig.~\ref{fig: showcase}).

\input{sub/method/fw}

\section{\texttt{TrIVD}: Tracking \& Image-Video Detection}
\label{sec: method}

In this section, we introduce \texttt{TrIVD}, the unified tracking and image-video object detection framework. In Sec.~\ref{sec: dynhead}, we first describe our unified representation for image and video inputs, and then propose the temporal-aware attention mechanism as an efficient spatio-temporal feature aggregation module for video inputs. In Sec.~\ref{sec: vl}, we introduce our unified classifier via region-text alignment for multi-task multi-dataset co-training, which handles the discrepancies and semantic overlaps of object categories across datasets. In Sec.~\ref{sec: det_track}, we detail and conclude \texttt{TrIVD}'s entire framework for unified tracking and image-video detection.


\subsection{\hspace*{-0.05cm}Temporal-Aware Unified Image-Video Backbone\hspace*{-0.2cm}}
\label{sec: dynhead}

We first introduce \texttt{TrIVD}'s unified formulation for image-video feature extraction in the backbone, then propose the temporal-aware attention module for video inputs.

\paragraph{Image/Video Inputs} 
\label{parag: input}
We represent each video as s list of frames and reshape its temporal dimension ($T$) into the batch dimension ($B$), to obtain a tensor $X \in \mathbb{R}^{B' \times H \times W \times C}$ with $B' = B*T$ as the new batch size, $H \times W$ refers to the spatial dimensions, and $C$ is the channel dimension. Similarly, we represent images as $X \in \mathbb{R}^{B \times H \times W \times C}$.



\paragraph{Backbone} 
\label{parag: backbone}
While our unified framework can use any vision
transformer architecture~\cite{dosovitskiy2020vit} to process the image and video inputs, we adopt the MViTv2~\cite{li2022mvit} architecture as the backbone, which hierarchically expands the feature complexity while reducing the spatial resolution via attention-pooling, given its proven advantage  
given its better performance and efficiency over single-scale vision transformers for image and video tasks~\cite{fan2021mvit,li2022mvit}. (See experimental results on the backbone gains (\texttt{TrIVD}$_{\text{backbone}}$) in Sec.~\ref{sec: compare}.)

With unified input formulation, the backbone maps input 2D patches into a \textit{shared} representation 
for images and videos, with a 2D linear layer 
followed by LayerNorm~\cite{ba2016ln}. 
Same embedding layers are applied to embed input (image/video) patches to enable maximal parameter sharing across visual modalities. Since all inputs are treated as single-frame images, only relative positional encoding~\cite{shaw2018respos} on the spatial domain is needed for either images or videos. 

Therefore, the frame-level multi-scale features extracted by MViTv2~\cite{li2022mvit} are a set of 3D features, 
\begin{equation}
\overline{\mathbf{F}} = \{\overline{F_l} ~\vert~ \overline{F_l} \in \mathbb{R}^{B' \times H_l \times W_l \times C_l},\, l = 1,\, ...,\, L\},
\end{equation}
where $H_l \times W_l$ refers to the spatial resolution at scale $l$, and $L$ denotes the number of spatial scales. Features of video inputs are then re-shaped back to their original dimensions:
\begin{equation}
\mathbf{F} ~=~ \{F_l \vert F_l \in \mathbb{R}^{B \times T \times H_l \times W_l \times C_l},\, l = 1,\, ...,\, L\},
\label{eq: reshaped}
\end{equation}
where $B,\, T$ denote the actual batch size and temporal length of the inputs respectively ($T \equiv 1$ for image inputs).

Our unified input formulation shares similar spirits with Omnivore~\cite{girdhar2022omnivore}, however, Omnivore represents both image and video inputs as \textit{videos}, i.e., as batches of \textit{4D} tensors $X \in \mathbb{R}^{B \times T \times H \times W \times C}$, where the temporal length ($T$) for image inputs are set to $1$. This results in 3D operations (e.g., 3D convolutions) and more expensive computations, especially for video OD/MOT problems that require high-resolution inputs. Instead, \texttt{TrIVD} treats video inputs as batches of \textit{images}, and conducts frame-level feature extraction first. Spatial features extracted in the backbone are then forwarded to our temporal-aware sequential attentions as described below, for spatio-temporal feature fusion.\footnote{We considered Omnivore~\cite{girdhar2022omnivore}'s video formulation when designing \texttt{TrIVD}'s unified backbone, yet found the proposed frame-level extraction with temporal-aware attention works better
(See comparisons in Sec.~\ref{sec: omnivore}.)}

\paragraph{Temporal-aware Sequential Attentions}
\label{parag: dynhead}

A major challenge of spatio-temporal feature aggregation is the trade-off between performance and computational cost. Dai et al.~\cite{dai2021dynhead} propose dynamic head for image OD, which decomposes attention on individual feature channels and improves the model's efficiency. We further extend this sequential attention idea to the \textit{temporal} dimension of videos. Specifically, given a set of multi-scale features $\mathbf{F}$ (Eq.~(\ref{eq: reshaped})), we decompose the overall attention function $\pi$ over the space-time domain, i.e., $W(\mathbf{F}) = \pi(\mathbf{F}) \cdot \mathbf{F}$, is decomposed into two sequential attentions along spatial and temporal axes:
\begin{equation}
    W(F_l)= \pi_{T_l} \big( \pi_{H_lW_l} (F_l) \big) \cdot F_l, \quad l = 1,\, ...,\, L,
    \label{eq: dynhead}
\end{equation}
where $\pi_{T_l}(\cdot),\, \pi_{H_lW_l}(\cdot)$ are two attention functions applied on the temporal axis ($T$), and spatial axis ($H_l\times W_l$) respectively (Fig.~\ref{fig: fw}). We follow~\cite{dai2021dynhead} for its attention design on the spatial domain ($\pi_{H_lW_l}$), and apply an additional temporal attention module by dynamically aggregating features along their temporal dimensions: 
\begin{equation}
    \pi_{T_l} (F_l) \cdot F_l = \frac{1}{T} \cdot \sigma \Big( f \big( \frac{1}{H_lW_l} \sum_{H_l,W_l}F_l \big) \Big), 
\label{eq: temp_attn}
\end{equation}
where $f(\cdot)$ is a linear function approximated by a $1 \times 1$ convolutional layer and $\sigma$ is the hard-sigmoid function.

The two sequential attention modules across spatial and temporal axes efficiently aggregate the spatio-temporal features from video inputs (Tab.~\ref{tab: omnivore}), and also achieve unified image-video representation. The extracted features could be further forwarded to \textit{any} downstream task-specific model.


\subsection{Unified Cross-dataset Classifier via Grounding}
\label{sec: vl}


One major task of detection/tracking is the class prediction for each proposed bounding box indicating the detected object category. It is typically achieved with a linear activation following the visual features of  extracted bounding boxes, and is trained with the multi-class cross entropy loss or focal loss~\cite{Lin2020FocalLF}. However, the above classification losses defined on logit-encoded class labels are not easily generalizable in the case of multi-dataset co-training --- Usually, annotated object class labels vary across different OD/MOT datasets, yet among them semantic overlaps may exist. 

One workaround is training with multiple dataset-specific classification layers~\cite{girdhar2022omnivore}, but this could potentially result in conflicts due to the \textit{non-exhaustive} annotations across datasets. A more elegant solution is binary classification with sigmoid activation~\cite{zhou2021detic}, where the judgement for every object category is independent from others. When the ground-truth categories are from other datasets, the related logits are simply masked out during gradient backpropagation. Yet this strategy still requires \textit{re-arranging} the class labels each time a new dataset is added to the training, which is not easily \textit{generalizable} to further explorations such as open-vocabulary settings. 

Aiming for a more general and flexible approach handling the mixed object categories and their semantic overlaps, \texttt{TrIVD} re-formulates the cross-dataset classification as phrase grounding~\cite{li2021glip,Zhang2022GLIPv2UL}, i.e., instead of classifying within $C$ fixed categories, we align proposed regions to classes represented in free form text. 
Specifically, during co-training, for each sample, we concatenate all available object categories in its belonged dataset to a text prompt. For instance, \texttt{VID}~\cite{russakovsky2015vid} dataset has the label to classname correspondences 
$C_{\texttt{VID}} = \{1: \text{airplane},\, ... ,\, 30: \text{zebra}\},
$
then the text prompt associated with \texttt{VID}'s samples is
$T_{\texttt{VID}} = ``\text{airplane} ~ ... ~ \text{zebra}\text{''},
$
where each class is converted to a word to be grounded, parsed by blank spaces. Thus unlike typical classification layers in detection/tracking models, \texttt{TrIVD} does \textit{not} directly output a class label for each proposed object, but encodes the text prompt with a pre-trained language model~\cite{Liu2019RoBERTaAR} and assesses the alignments between the visual and text representations~(Fig.~\ref{fig: fw}).

Following~\cite{Oord2018RepresentationLW,li2021glip,kamath2021mdetr}, we replace the regular cross entropy loss with soft token loss ($\mathcal{L}_{\text{soft}}$) which encourages the predicted token spans to be aligned with the objects' semantic meanings, and the contrastive alignment loss ($\mathcal{L}_{\text{contrast}}$), which increases similarities between the embedded representations of the detected objects and the matched words in text. 

\input{sub/exp/fig/mot_comp_fig}

\subsection{Unified Tracking and Image-Video Detection}
\label{sec: det_track}

As illustrated in Fig.~\ref{fig: fw}, the proposed \texttt{TrIVD} consists of two major components: 1) Modality-agnostic visual feature extraction in the backbone, where one could opt to conduct frame-level feature extraction for images, or add spatial-temporal feature fusion for video inputs with the introduced temporal-aware attention module (Sec.~\ref{parag: dynhead}); 2) Self-attention-based unified detector-tracker as follows. 

\paragraph{DETRs}
Our detector/tracker is built upon the end-to-end detection framework, Deformable DETR~\cite{carion2020detr,zhu2020deformable}, 
as its self-attention mechanism could be simultaneously adopted for object detection as well as tracking's data association (Sec.~\ref{parag: track}). Briefly speaking, with a transformer encoder-decoder structure~\cite{vaswani2017attn}, Deformable DETR is initialized with a certain number ($N_{\text{box}}$) of empty bounding boxes (i.e., object queries), to detect potentially existing objects in the boxes. Forwarding through the cross-attention modules in the transformer's decoder, \texttt{TrIVD} outputs the final predictions of the box coordinates along with corresponding class label and confidence score. Deformable DETR is trained with the Hungarian matching loss, where a bipartite matching is computed between the $N_{\text{box}}$ predicted object queries and the ground-truth objects. The matched objects are encouraged to align with the ground-truth, while the un-matched ones are treated as background. Cross-entropy loss is used for classification supervision, $L_1$ loss and Generalized IoU~\cite{Rezatofighi2019GeneralizedIO} are used for the bounding box supervision.

\paragraph{Detection-Tracking Bipartite Matching} 
\label{parag: track}

Since \texttt{TrIVD} does not directly predict class labels, but aligns the token positions in the text prompt with the proposed objects~(Sec.~\ref{sec: vl}), 
the bipartite matching between the ground truth and proposed objects do \textit{not} rely on class labels, but on the \textit{relevant} positions of the classname in the \hypertarget{sec: det}{text prompt}. 

\noindent \textbf{- Detection} only cares about the proposed objects of the \textit{current} frame. Therefore, in our unified detection-tracking context, we treat all detected objects as \textit{newly appeared}. Thus the bipartite matching happens between the proposed and the ground truth objects, same as in DETRs~\cite{carion2020detr,zhu2020deformable}.

\noindent \textbf{- Tracking}, in addition to localization and classification of objects in the current frame, requires the knowledge of object/track identities \textit{across} video frames, and faces the challenges of objects disappearing from or re-entering the scene. Thanks to the self-attention mechanism in transformers~\cite{vaswani2017attn} which correlates all components across the entire inputs, data association across video frames could be achieved in a detection/tracking-by-attention fashion~\cite{carion2020detr,zhu2020deformable,meinhardt2021trackformer}. Specifically, the frame-to-frame data association is realized by 1) integrating previous frame's features into current frame's transformer encoder, where a temporal feature encoding~\cite{Wang2021EndtoEndVI} is used to enable queries to discriminate between features from the previous frame; and 2) adding the previous detected object queries, named track queries, to the initialization of new object queries for the current frame, and together forward into the transformer's decoder of current frame~(Fig.~\ref{fig: fw}). In the transformer decoder, computing self-attention between adjacent frame features as well as between newly initialized object queries and track queries, naturally performs the detection of new objects while avoiding re-detection of already detected/tracked objects~\cite{meinhardt2021trackformer}.

Therefore, the bipartite matching for tracking contains two scenarios. 1) If the objects in the current frame have already presented in previous frame(s), the mapping depends on the ground truth track \textit{identities}~\cite{meinhardt2021trackformer}; 2) Otherwise, the mappings to newly-appeared objects or background reduce to the same matching plan as described in \hyperlink{sec: det}{\textbf{Detection}}. 

In summary, the bipartite matching loss, for either detection or tracking, is achieved by solving a minimum cost assignment problem~\cite{carion2020detr,Stewart2015EndtoEndPD}, resulting in the following combined end-to-end training loss for \texttt{TrIVD}:
\begin{equation}
    \mathcal{L} = \mathcal{L}_{\text{soft}} + \mathcal{L}_{\text{contrast}} + \mathcal{L}_{\text{box\_detect}} + \mathcal{L}_{\text{box\_track}} \, ,
    \label{eq: loss}
\end{equation}
where $\mathcal{L}_{\text{soft}},\, \mathcal{L}_{\text{contrast}}$ are the object category prediction losses~(Sec.~\ref{sec: vl}), $L_1$ loss and Generalized IoU~\cite{Rezatofighi2019GeneralizedIO} are used as the box prediction losses for both tracking object boxes ($\mathcal{L}_{\text{box\_track}}$) and newly-appeared/non-object boxes  
($\mathcal{L}_{\text{box\_detect}}$). Since for detection tasks, we treat all proposed objects as new detections, thus $\mathcal{L}_{\text{box\_track}} \equiv 0$. 

\section{Experiments}
\label{sec: exp}


\subsection{Datasets and Metrics}
\label{sec: datasets}

\paragraph{Image and Video Object Detection (OD)} 
~ \vspace{0.05cm} \\
\noindent \textbf{- COCO}~\cite{lin2014coco} is an image OD dataset with 80 annotated categories. All models in this paper are trained on its 118K training images and evaluated on its 5K validation images. 

\noindent \textbf{- VID}~\cite{russakovsky2015vid} is a video OD dataset, containing 3862 training and 555 validation videos. \texttt{VID} has 30 annotated categories, among which 13 categories overlap with those in \texttt{COCO}. We also follow the previous video OD work~\cite{Deng2019RelationDN,Wang2018FullyMN,zhu2017dff,Yao2020VideoOD} and include \texttt{DET} ~\cite{Russakovsky2015ilsvrc} dataset in the training set.

\noindent \textbf{- Metrics}~ For image OD, we use the official metrics on average precision (AP) from \texttt{COCO}~\cite{lin2014coco}, i.e., AP, AP$_{50}$, AP$_{75}$, AP$_{\text{S}}$, AP$_{\text{M}}$, and AP$_{\text{L}}$. For video OD, AP is used as the evaluation metric following previous work~\cite{Deng2019RelationDN,Wang2018FullyMN,zhu2017dff,Yao2020VideoOD}.

\paragraph{Multi-object Tracking (MOT)} 
~ \vspace{0.05cm} \\
\noindent \textbf{- MOT17}~\cite{Milan2016MOT16AB} is a person-annotated tracking benchmark, with 7 sequences in train and test sets respectively.


\noindent \textbf{- Metrics}~ Varying metrics are used to evaluate different aspects of MOT performance~\cite{Bernardin2008EvaluatingMO,meinhardt2021trackformer}. We adopt the 7 widely-used metrics~\cite{Ristani2016PerformanceMA,Milan2016MOT16AB}: multiple object tracking accuracy (MOTA), identity F1 score (IDF1), mostly tracked (MT), mostly lost (ML), false positive (FP) and false negative (FN), and number of identity
switches (IDS). 

\paragraph{Zero-shot Tracking Evaluation} 
~ \vspace{0.05cm} \\
\noindent \textbf{- TAO}~\cite{Dave2020TAOAL} is a large-scale dataset of videos with labeled object tracks. To evaluate \texttt{TrIVD}'s zero-shot tracking ability, we directly \textit{test} \texttt{TrIVD}, which has been co-trained across \texttt{COCO}, \texttt{VID} and \texttt{MOT17}, on \texttt{TAO}'s annotated tracking ground truth on \texttt{VID}. Note that \texttt{TAO}'s tracking annotations of \texttt{VID} are \textit{not} used during \texttt{TrIVD}'s training.

\noindent \textbf{- Metrics}~ We use the same metrics, Det mAP and Track mAP, as in \texttt{TAO}~\cite{Dave2020TAOAL} for zero-tracking evaluation (Sec.~\ref{sec: quantitative}).

\subsection{Implementation Details}
\label{sec: implement}

We use MViTv2-s~\cite{fan2021mvit,li2022mvit} as the backbone, we follow deformable DETR~\cite{zhu2020deformable} for its end-to-end transformer-based structure, and TrackFormer~\cite{meinhardt2021trackformer} for its track queries aggregation and augmentations. To make sure we can cover  
objects in the crowded scenes in \texttt{MOT17}~\cite{Milan2016MOT16AB} tracking dataset, we set the number of object queries to $N_{\text{box}} = 500$. For the MViTv2-s backbone, we follow~\cite{li2022mvit} and pre-train the backbone on \texttt{ImageNet-21K}~\cite{Deng2009ImageNetAL}  and fine-tune on \texttt{COCO} with 36 epochs. Our training schedules follow~\cite{zhu2020deformable}, and we set the batch size as 2 with initial learning rates of 0.0001 for deformable DETR encoder-decoder, and 0.00001 for the backbone. For the language model, we follow~\cite{kamath2021mdetr} and use the HuggingFace's~\cite{Wolf2020TransformersSN} pre-trained RoBERTa-base~\cite{Liu2019RoBERTaAR} as our text encoder. We use a linear decay with warm-up schedule, increasing linearly to 0.00005 during the first 1\% of the total number of steps, then decreasing linearly back to 0 for the rest of the training.


\subsection{Benchmark Results: Image/Video OD, MOT}
\label{sec: compare}

\begin{table}[t]
\hspace*{-0.25cm} 
\resizebox{1.05\linewidth}{!}{
\centering 
    \begin{tabular}{lcccccccc}
       \toprule \\ [-3.3ex] 
      \multicolumn{1}{l}{\multirow{2}{*}{\bf Method}} &
      \multicolumn{1}{c}{\multirow{2}{*}{\bf Backbone}} &
      \multicolumn{1}{c}{\multirow{2}{*}{\bf Detector}} &
      \multicolumn{1}{c}{\multirow{2}{*}{\bf AP~$\uparrow$}} & \multicolumn{1}{c}{\multirow{2}{*}{\bf AP$_{\text{50}}$~$\uparrow$}} & \multicolumn{1}{c}{\multirow{2}{*}{\bf AP$_{\text{75}}$~$\uparrow$}} & \multicolumn{1}{c}{\multirow{2}{*}{\bf AP$_{\text{S}}$~$\uparrow$}} & \multicolumn{1}{c}{\multirow{2}{*}{\bf AP$_{\text{M}}$~$\uparrow$}} & \multicolumn{1}{c}{\multirow{2}{*}{\bf AP$_{\text{L}}$~$\uparrow$}} \\ [1.7ex]
        
    
     \midrule
     
      Faster-RCNN~\cite{Girshick2015frcnn} & ResNet-50 & Faster-RCNN & 42.0 & 62.1 & 45.5 & 26.6 & 45.4 & 53.4 \\ [0.1ex]
      DETR~\cite{carion2020detr} & ResNet-50 & DETR & 42.0 & 62.4 & 44.2 & 20.5 & 45.8 & 61.1 \\ [0.1ex]
      DETR-DC5~\cite{carion2020detr} & ResNet-50 & DETR & 43.3 & 63.1 & 45.9 & 22.5 & 47.3 & 61.1 \\ [0.1ex]
      Def-DETR~\cite{zhu2020deformable} & ResNet-50 & Def-DETR & 43.8 & 62.6 & 47.7 & 26.4 & 47.1 & 58.0 \\ [0.1ex]
      Def-DETR$^{\text{box-refine}}$~\cite{zhu2020deformable} & ResNet-50 & Def-DETR & 45.4 & 64.7 & 49.0 & 26.8 & 48.3 & 61.7 \\ [0.1ex]
      
     \hline\\[-2.ex]
     
      \texttt{TrIVD$_{\text{single}}$} & MViTv2-s & Def-DETR & 46.2 & 65.1 & 48.9 & \textbf{27.9} & 48.9 & 61.6 \\ [0.1ex]
      \texttt{TrIVD$_{\text{multi}}$} & MViTv2-s & Def-DETR & \textbf{46.5} & \textbf{65.7} & \textbf{49.3} & 27.5 & \textbf{48.9} & \textbf{61.9} \\ [0.1ex]

\bottomrule  
	\vspace{-0.7cm}
    \end{tabular}  
    \caption{Comparisons between \texttt{TrIVD} and the state-of-the-art image OD approaches on \texttt{COCO} 2017 validation set. Deformable-DETR (Def-DETR)~\cite{zhu2020deformable} could be viewed as our plain baseline on image OD: \texttt{TrIVD$_{\text{single}}$} equals Def-DETR when we switch the MViTv2-s~\cite{li2022mvit} backbone to ResNet-50~\cite{He2016DeepRL}.} 
    \label{tab: coco}
}

\end{table}

\begin{table}[t]
\resizebox{0.85\linewidth}{!}{
\centering 
    \begin{tabular}{lccccc}
       \toprule \\ [-3.4ex] 
      \multicolumn{1}{l}{\multirow{2}{*}{\bf Method}} &
      \multicolumn{1}{c}{\multirow{2}{*}{\bf Backbone}} &
      \multicolumn{1}{c}{\multirow{2}{*}{\bf Detector}}
      & \multicolumn{1}{c}{\multirow{2}{*}{$\mathbf{N_{\text{frame}}}$}}
      & \multicolumn{1}{c}{\multirow{2}{*}{\bf AP~$\uparrow$}}
      \\ [1.7ex]
        
        
     \midrule
     
      DFF~\cite{zhu2017dff} & ResNet-50 & Faster-RCNN & 10 & 70.4 \\ 
      FGFA~\cite{zhu2017fgfa}  & ResNet-50 & Faster-RCNN & 21 & 74.0 \\ 
      RDN~\cite{Deng2019RelationDN}  & ResNet-50 & Faster-RCNN & 3 & 76.7 \\ 
      MEGA~\cite{chen2020mega}  & ResNet-50 & Faster-RCNN & 9 & 77.3 \\ 
      TransVOD~\cite{Yao2020VideoOD} & ResNet-50 & Def-DETR & 3 & 77.7 \\ 
      Def-DETR~\cite{zhu2020deformable} & ResNet-50 & Def-DETR & 1 & 76.0 \\ 
      
     \hline\\[-2.ex]
     
      \texttt{TrIVD$_{\text{single}}$} & MViTv2-s & Def-DETR  & 3 & 77.9  \\ 
      \texttt{TrIVD$_{\text{multi}}$} & MViTv2-s & Def-DETR  & 3 & \textbf{78.3}  \\ 

\bottomrule  
	\vspace{-0.7cm}
    \end{tabular}  
    \caption{Comparisons between \texttt{TrIVD} and the state-of-the-art video OD approaches on \texttt{VID} validation set. $\mathbf{N_{\text{frame}}}$ refers to the temporal length of corresponding models' input video clips. Deformable-DETR (Def-DETR)~\cite{zhu2020deformable} could be viewed as our per-frame detection baseline on video OD: \texttt{TrIVD$_{\text{single}}$} reduces to Def-DETR when we switch the MViTv2-s~\cite{li2022mvit} backbone to ResNet-50~\cite{He2016DeepRL} and perform frame-by-frame detection for \texttt{VID}, i.e., without temporal-aware attention.} 
    \label{tab: vid}
}

\end{table}

\begin{table}[t]
\hspace*{-0.25cm} 
\resizebox{1.05\linewidth}{!}{
\centering 
    \begin{tabular}{lccccccccc}
       \toprule \\ [-3.5ex] 
      \multicolumn{1}{l}{\multirow{2}{*}{\bf Method}} &
      \multicolumn{1}{c}{\multirow{2}{*}{\bf Data}} & 
      \multicolumn{1}{c}{\multirow{2}{*}{\bf Backbone}} & 
      \multicolumn{1}{c}{\multirow{2}{*}{\bf MOTA~$\uparrow$}} & \multicolumn{1}{c}{\multirow{2}{*}{\bf IDF1~$\uparrow$}} & \multicolumn{1}{c}{\multirow{2}{*}{\bf MT~$\uparrow$}} & \multicolumn{1}{c}{\multirow{2}{*}{\bf ML~$\downarrow$}} & \multicolumn{1}{c}{\multirow{2}{*}{\bf FP~$\downarrow$}} & \multicolumn{1}{c}{\multirow{2}{*}{\bf FN~$\downarrow$}} & \multicolumn{1}{c}{\multirow{2}{*}{\bf IDS~$\downarrow$}} \\ [1.7ex]
        

     \midrule 
     
      FAMNet~\cite{Chu2019FAMNetJL} & - & ResNet-101 & 52.0 & 48.7 & 450 & 787 & 14138 & 253616 & 3072 \\ 
      Tracktor++~\cite{Bergmann2019TrackingWB} & M \& C & ResNet-101 & 56.3 & 55.1 & 498 & 831 & \textbf{8866} & 235449 & 1987 \\ 
      GSM~\cite{Liu2020GSMGS} & M \& C & ResNet-34 & 56.4 & 57.8 & 523 & 813 & 14379 & 230174 & \textbf{1485} \\ 
      CenterTrack~\cite{Zhou2020TrackingOA} & - & DLA~\cite{Yu2018DeepLA} & 60.5 & 55.7 & 580 & 777 & 11599 & 208577 & 2540 \\ 
      TMOH~\cite{Stadler2021ImprovingMP} & - & ResNet-101 & 62.1 & \textbf{62.8} & 633 & 739 & 10951 & 201195 & 1897 \\ 
      TrackFormer~\cite{meinhardt2021trackformer} & - & ResNet-50 & 62.3 & 57.6 & 688 & 638 & 16591 & 192123 & 4018 \\ 
      
     \hline\\[-2.ex]
     
      \texttt{TrIVD$_{\text{single}}$} & - & MViTv2-s & 62.5 & 58.1 & 671 & 613 & 15896 & 190325 & 4072 \\ 
      \texttt{TrIVD$_{\text{multi}}$} & - & MViTv2-s & \textbf{64.8} & 60.1 & \textbf{724} & \textbf{598} & 15332 & \textbf{187232} & 3967 \\ 

\bottomrule  
	\vspace{-0.7cm}
    \end{tabular}  
    \caption{Comparisons between \texttt{TrIVD} and the state-of-the-art MOT approaches on \texttt{MOT17} test set (online public detections results reported). The 2$^{\text{nd}}$ column indicates extra tracking data included in the training (M: Market1501~\cite{Zheng2015ScalablePR}; C: CUHK03~\cite{Li2014DeepReIDDF}).} 
    \label{tab: mot}
}

\end{table}


We examine \texttt{TrIVD}'s performance on \texttt{COCO}~\cite{lin2014coco}, \texttt{VID}~\cite{russakovsky2015vid} and \texttt{MOT17}~\cite{Milan2016MOT16AB} datasets. To better understand \texttt{TrIVD}'s benefits, we implement two setups for \texttt{TrIVD}, focusing on the gains from different parts of \texttt{TrIVD}: 

\noindent 1) \texttt{TrIVD$_{\text{backbone}}$} - the \texttt{TrIVD} model, \textit{but} trained on \textit{individual} datasets in a single-task manner. \texttt{TrIVD$_{\text{backbone}}$} is used to demonstrate the benefits from our unified image-video backbone feature fusion (Sec.~\ref{sec: dynhead}). \hypertarget{model: multitask}{~}

\noindent 2) \texttt{TrIVD$_{\text{multitask}}$} ($\equiv$\texttt{TrIVD}) - the full \texttt{TrIVD} model and setup, i.e., co-trained on \texttt{COCO} \& \texttt{VID} \& \texttt{MOT17} in a multi-dataset, multi-task manner. \texttt{TrIVD$_{\text{multitask}}$} is used to demonstrate the benefits from our unified detector-tracker design and multi-task co-training (Sec.~\ref{sec: det_track}). 

Tabs.~\ref{tab: coco}-\ref{tab: mot} compare \texttt{TrIVD$_{\text{multitask}}$}'s performance with single-task baselines on all image OD, video OD and MOT tasks. The proposed \texttt{TrIVD$_{\text{multitask}}$} achieves better performances across all evaluation metrics on \texttt{COCO}~\cite{lin2014coco}, especially on small objects (AP$_{\text{S}}$)~(Tab.~\ref{tab: coco}). Comparisons on \texttt{VID}~\cite{russakovsky2015vid} dataset demonstrate the effectiveness of the proposed temporal-aware attention module in aggregating spatio-temporal features of video inputs~(Tab.~\ref{tab: vid}). We also achieve better MOT performance on MOTA, MT, ML as well as FN, \textit{without} the need for training on additional tracking data as used in ~\cite{Bergmann2019TrackingWB,Liu2020GSMGS}~(Tab.~\ref{tab: mot}). 

Further, comparisons between the single-task trained model (\texttt{TrIVD$_{\text{backbone}}$}) and the full multi-dataset multi-task co-trained model (\texttt{TrIVD$_{\text{multitask}}$}) explicitly highlights the effectiveness of multi-task co-training strategy achieved with our proposed unified formulation -- \texttt{TrIVD$_{\text{multitask}}$} outperforms \texttt{TrIVD$_{\text{backbone}}$} over image OD, video OD and MOT tasks, especially for MOT where we observe an improvement on MOTA by 3.7\%~(Tab.~\ref{tab: mot}).  


\subsection{A New Task: Zero-shot Tracking}
\label{sec: zero-shot}

With the unified formulation and multi-task co-training, \texttt{TrIVD} further achieves \textit{zero-shot} tracking, on object categories which do not exist in the tracking training dataset. 
 

\paragraph{Qualitative Visualizations}
\label{sec: qualitative}

\input{sub/exp/fig/det_all_fig}

\texttt{TrIVD}'s unified formulation allows us to conduct image OD, video OD and MOT within one model (Fig.~\ref{fig: showcase}), and further extend tracking to a wider range of object categories, achieving zero-shot tracking capability. As shown in Fig.~\ref{fig: mot}, designed and trained specifically on \texttt{MOT17}, a person-tracking dataset, a typical tracking model such as TrackFormer~\cite{meinhardt2021trackformer} can detect and track people identities (Fig.~\ref{fig: mot}, $1^{\text{st}}$ column), but it is not able to track other object categories that are not annotated by \texttt{MOT17}, e.g., cars, birds, pandas. In contrast, with our unified formulation, \texttt{TrIVD} that is co-trained on OD datasets (\texttt{COCO}, \texttt{VID}) is now able to borrow its knowledge learned from the detection data and achieve zero-shot tracking on novel objects without training on their tracking annotations. We can therefore track both the people and the cars in the same street view from \texttt{MOT17} (Fig.~\ref{fig: mot}, $2^\text{nd}$--$3^\text{rd}$ columns).

We further test \texttt{TrIVD}'s zero-shot tracking ability on \texttt{VID} (video OD), where no ground truth tracking annotation is available (Fig.~\ref{fig: showcase}, $2^\text{nd}$--$4^\text{th}$ columns; Fig.~\ref{fig: mot}, $4^\text{th}$--$7^\text{th}$ columns). \texttt{TrIVD} detects and tracks the objects with their \textit{position} changes successfully (e.g., airplanes and pandas in Fig.~\ref{fig: showcase}, motorcycles in Fig.~\ref{fig: mot}), and handles object \textit{disappearing} scenarios well (e.g., cars in Fig.~\ref{fig: showcase}, airplanes in Fig.~\ref{fig: mot}). Another challenge for MOT is identifying previous objects that \textit{re-enter} the scene, we indeed observe some failures. In Fig.~\ref{fig: mot}, $6^\text{th}$ column, \texttt{TrIVD} re-identifies the bird (pink) when it re-enters the scene, but fails to recognize the same bicycle (Fig.~\ref{fig: mot}, $7^\text{th}$ column) under significant camera view and pose changes, and identifies it as a new bicycle (blue $\rightarrow$ green).


\paragraph{Quantitative Evaluation}
\label{sec: quantitative}


\begin{table}[t]
\resizebox{\linewidth}{!}{
\centering 
    \begin{tabular}{cccc}
       \toprule \\ [-3.1ex] 
      {\bf Model} & \multicolumn{1}{c}{\multirow{1}{*}{\bf Testing Category}} &
      \multicolumn{1}{c}{\multirow{1}{*}{\bf Detect mAP~$\uparrow$}} & \multicolumn{1}{c}{\multirow{1}{*}{\bf Track mAP~$\uparrow$}} \\ [-0.6ex]

     \midrule\\[-2.5ex]
      TrackFormer~\cite{meinhardt2021trackformer} & person (\textit{\color{blue}known}) & 71.8 & 32.6 \\ [0.1ex]
      \texttt{TrIVD} & person (\textit{\color{blue}known}) & \textbf{72.3} & \textbf{33.1}  \\ [0.2ex]
      \hline \\[-2.2ex]
      \hline \\[-2.2ex]
      \texttt{TAO} reported & non-person (\textit{\color{blue}known}) & 79.2 & 60.3 \\ [0.2ex]
      \hline \\[-2.2ex]
      TrackFormer~\cite{meinhardt2021trackformer} & non-person (\textit{\color{red}unknown}) & 0 & 0 \\ [0.2ex]
      \texttt{TrIVD} & non-person (\textit{\color{red}unknown}) & \textbf{77.9} & \textbf{43.6} \\ [-0.4ex]
\bottomrule  \\ [-2.1ex]
\multicolumn{4}{l}{\multirow{1}{*}{\textit{\color{blue}known}: testing category included in ground truth tracking data during training;}}  \\
\multicolumn{4}{l}{\multirow{1}{*}{\textit{\color{red}unknown}: testing category \textit{not} in ground truth tracking data during training.}}  \\
\bottomrule

    \end{tabular}
    
      \caption{\texttt{TrIVD}'s zero-shot (\textit{\color{red}unknown}) tracking on \texttt{TAO} compared with a regular person-tracker, TrackFormer~\cite{meinhardt2021trackformer}.}
    \label{tab: tao}
}
\end{table}

As the first zero-shot tracking model, \texttt{TrIVD} does \textit{not} have comparable baselines. We \textit{directly} test  \texttt{TrIVD$_{\text{multitask}}$} (from Sec.~\hyperlink{model: multitask}{4.3}) on \texttt{TAO}~\cite{Dave2020TAOAL}'s annotated person and \texttt{VID} (non-person) tracking data (Tab.~\ref{tab: tao}). A regular tracker trained on person-annotated tracking data cannot detect/track any non-person object (4$^{\text{th}}$ row), yet \texttt{TrIVD$_{\text{multitask}}$}, co-trained solely with their detection annotations, achieves decent tracking performance, and sets the \textit{first} zero-shot tracking baseline. 


\subsection{Ablations and Further Discussions}
\label{sec: ablations}
We explored the benefits of \texttt{TrIVD}'s unified backbone and multi-task co-training in Sec.~\ref{sec: compare}. Here, we conduct ablations on 1) the frame-level feature extraction with temporal-aware attention in \texttt{TrIVD}'s backbone (Sec.~\ref{parag: backbone}); 2) the temporal feature aggregation of \texttt{TrIVD} (Sec.~\ref{parag: dynhead}) for video OD. We also discuss \texttt{TrIVD}'s broader impacts and future directions. 


\setlength\intextsep{0pt}
\begin{table}[t]


\resizebox{\linewidth}{!}{
\centering 
    \begin{tabular}{ccccccc}
       \toprule \\ [-3.1ex] 
      \multicolumn{1}{c}{\multirow{2}{*}{\bf Feature Fusion}} &
      \multicolumn{1}{c}{\multirow{2}{*}{\bf Training Data}} & \multicolumn{3}{c}{\multirow{1}{*}{\bf \texttt{COCO}}} & {\bf \texttt{VID}} & \multicolumn{1}{c}{\multirow{2}{*}{\bf Avg. Time ($s$) / Iter.~$\downarrow$}} \\ [-0.85ex]
      \cmidrule(lr){3-5} \\ [-3.1ex]
       & &
      \multicolumn{1}{c}{\multirow{1}{*}{\bf AP~$\uparrow$}} & \multicolumn{1}{c}{\multirow{1}{*}{\bf AP$_{\text{50}}$~$\uparrow$}} & \multicolumn{1}{c}{\multirow{1}{*}{\bf AP$_{\text{75}}$~$\uparrow$}} & \multicolumn{1}{c}{\multirow{1}{*}{\bf AP$_{\text{S}}$~$\uparrow$}} \\ [-0.8ex]
        
    
     \midrule
     
      3D-Conv (Omnivore~\cite{girdhar2022omnivore}) & \texttt{COCO} & 45.7 & 63.9 & 48.1 & - & 1.49\\ [0.ex]
      2D-Conv+Attn (\texttt{TrIVD}) & \texttt{COCO} & {\bf{46.2}} & {\bf{65.1}} & {\bf{48.9}} & - & {\bf{1.36}} \\ [0.ex]
      \hline \\[-2.1ex]
      3D-Conv (Omnivore~\cite{girdhar2022omnivore}) & \texttt{VID} & - & - & - & 77.1 & 1.81 \\ [0.ex]
      2D-Conv+Attn (\texttt{TrIVD}) & \texttt{VID} & - & - & - & {\bf{77.9}} & {\bf{1.62}} \\ [0.ex]
      \hline \\[-2ex]
      3D-Conv (Omnivore~\cite{girdhar2022omnivore}) & \texttt{COCO} + \texttt{VID} & 46.0 & 64.9 & 48.8 & 78.0 & 1.63\\ [0.ex]
      2D-Conv+Attn (\texttt{TrIVD}) & \texttt{COCO} + \texttt{VID} & {\bf{46.5}} & {\bf{65.7}} & {\bf{49.3}} & {\bf{78.3}} & {\bf{1.45}} \\ [-0.4ex]

\bottomrule  
    
    \end{tabular}
    \caption{Ablations on image-video feature fusions in the backbone. All models use MViTv2-s~\cite{li2022mvit} as feature extraction, and Def-DETR~\cite{carion2020detr} as detector.}
    \label{tab: omnivore}
}

\end{table}


\paragraph{Unified Image-Video Backbone Fusion Designs}
\label{sec: omnivore}

Omnivore~\cite{girdhar2022omnivore} presents a 3D-Conv backbone to create a single embedding for multiple visual modalities, which is a parallel work of \texttt{TrIVD} yet we have different \textit{focuses}: Image-video feature fusion is part of \texttt{TrIVD}'s unified backbone designs. We explored 3D-Conv (Omnivore) and 2D-Conv+Attn (\texttt{TrIVD}, Sec.~\ref{sec: dynhead}) image-video backbone in our early-stage research, and found ours performs \textit{better} and \textit{faster} in image-video OD co-training task (See Tab.~\ref{tab: omnivore}).

\paragraph{Number of Reference Frames in Video OD}
\label{sec: n_ref}


 Without any temporal-aware feature fusion ($\mathbf{N_{\text{frame}}}$ as 1), \texttt{TrIVD} reduces to a per-frame image OD model. With only 2 reference frames aggregated in temporal fusion, we observe a significant boost on \texttt{TrIVD$_{\text{backbone}}$}'s performance (Tab.~\ref{tab: num_frame})

\begin{table}[t]
\resizebox{0.6\linewidth}{!}{
\centering 
    \begin{tabular}{ccccccc}
       \toprule \\ [-3.4ex] 
      \multicolumn{1}{l}{\multirow{2}{*}{ $\mathbf{N_{\text{frame}}}$}} &
      \multicolumn{1}{c}{\multirow{2}{*}{\bf 1}} &
      \multicolumn{1}{c}{\multirow{2}{*}{\bf 3}} & \multicolumn{1}{c}{\multirow{2}{*}{\bf 5}} & \multicolumn{1}{c}{\multirow{2}{*}{\bf 7}} & \multicolumn{1}{c}{\multirow{2}{*}{\bf 9}} & \multicolumn{1}{c}{\multirow{2}{*}{\bf 11}}
      \\ [1.7ex]
        
        
     \midrule
     
      \textbf{AP}~$\uparrow$ & 76.8 & 77.9 & 78.6 & \textbf{79.4} & 79.2 & 79.3 \\ 

\bottomrule  
	\vspace{-0.55cm}
    \end{tabular}  
    \caption{Ablations on the number of frames aggregated in an input video clip for video OD on \texttt{VID}~\cite{russakovsky2015vid} dataset (Model: \texttt{TrIVD$_{\text{single}}$}).} 
    \label{tab: num_frame}
}

\end{table}

 \noindent and $\mathbf{N_{\text{frame}}}$ as 7 for the input video clips achieves the best video OD performance. But continuing to increase the number of aggregated frames does not bring obvious gains further. 

\paragraph{Open-vocabulary Detection/Zero-shot Tracking}
\label{sec: open-vocab}

With unified grounding-based classifier~(Sec.~\ref{sec: vl}), we naturally extend the \texttt{TrIVD}'s detection ability to the combined annotated object categories of all co-trained datasets (\texttt{COCO}, \texttt{VID}, \texttt{MOT17}). Thus \texttt{TrIVD} detects other objects than people (e.g., cars, bicycles, motorcycles) in the person-annotated-only \texttt{MOT17} dataset (Fig.~\ref{fig: det}, $3^\text{rd}$ column). 
Based on region-text alignment, \texttt{TrIVD}'s detection/tracking vocabulary could be further scaled up upon being pre-trained on larger OD datasets such as Objects365~\cite{Shao2019Objects365AL}, or/and on semantic-rich phrase grounding datasets~\cite{li2021glip,Zhang2022GLIPv2UL}, e.g., Flickr30K~\cite{Plummer2015Flickr30kEC}, VG Caption~\cite{Krishna2016VisualGC}. Our on-going research focuses on \texttt{TrIVD}'s open-vocabulary image-video detection-tracking abilities. 

\section{Conclusion}
\label{sec: con}

We introduced \texttt{TrIVD} which performs image object detection, video object detection, and multi-object tracking within a unified framework. We unified image and video inputs with spatial-temporal-aware deep feature fusion, and connected image-video detection with tracking via self-attention. Our unified classifier based on region-text alignment naturally extends the detection/tracking vocabulary of \texttt{TrIVD} and enables zero-shot tracking. We hope this work brings deeper insights and reveals greater power of the multi-task learning for image-video object detection and multi-object tracking.



{\small
\bibliographystyle{ieee_fullname}
\bibliography{reference}
}


\newpage
\appendix

\clearpage

\setcounter{figure}{1}
\setcounter{equation}{1}

\renewcommand\thefigure{\thesection.\arabic{figure}}
\renewcommand\theequation{\thesection.\arabic{equation}}

{\centering 
\section*{Unifying Tracking and Image-Video Detection\\\vspace{0.05cm}(Appendix)}
}

\vspace{0.1cm}
\noindent This Appendix includes: 

\begin{itemize}

\item[{\bf{\ref{app: classifier}:}}] Full details on the grounding-based unified classifier;

\item[{\bf{\ref{app: detr}:}}] Additional details on the unified detector-tracker;

\item[{\bf{\ref{app: implement}:}}] More implementation details and training recipes;

\item[{\bf{\ref{app: discussion}:}}] Further discussions on \texttt{TrIVD};


\end{itemize}

We also provide the complete zero-shot multi-object tracking ``.mp4'' videos. Please kindly refer to the attached videos in the sub-folder ``\textit{./zero-shot-tracking}''.

\vspace{0.3cm}




\section{Additional Details on the Unified Cross-dataset Classifier}
\label{app: classifier}

As introduced in Sec.~\ref{sec: vl}, the proposed \texttt{TrIVD}'s unified cross-dataset classifier follows previous open-vocabulary detection work~\cite{li2021glip,Zhang2022GLIPv2UL,kamath2021mdetr}, and uses the corresponding grounding-based losses ($\mathcal{L}_{\text{soft}}$ \& $\mathcal{L}_{\text{contrast}}$). To make this paper more self-contained, we include the complete details of \texttt{TrIVD}'s classifier design here. See further experimental comparisons between traditional cross-entropy loss based classification and region-text alignment based classification in Appendix~\ref{app: cross-entropy}.

As fully end-to-end detection frameworks, the original DETRs~\cite{carion2020detr,zhu2020deformable} are trained for object localization and classification jointly. Briefly, during training, DETR computes a bipartite matching between the $N$ proposed objects and the ground-truths. For object those are matched, DETR is supervised by the corresponding ground truth targets. For object those are un-matched, DETR is enforced to predict the non-object label $\emptyset$. DETR's classification head is supervised using standard cross-entropy loss, while the bounding box head is supervised using a combination of absolute error (L1 loss) and Generalized IoU~\cite{carion2020detr}. 

\texttt{TrIVD} keeps the same bipartite matching idea as DETR, yet switching from the traditional cross-entropy classification loss to the two grounding-based losses ($\mathcal{L}_{\text{soft}}$ \& $\mathcal{L}_{\text{contrast}}$) following~\cite{Oord2018RepresentationLW,kamath2021mdetr}, to encourage alignment between the image and the text. Specifically, we first convert the digit labels of all datasets to corresponding free form text as the text prompt input of a language encoder. For example, \texttt{VID}~\cite{russakovsky2015vid} dataset has the following label to object category correspondences,
$$ C_{\texttt{VID}} = \{1: \text{airplane},\, 2: \text{antelope}, \, ... ,\, 30: \text{zebra}\} \,,
$$ 
then the converted input text prompt associated with samples from \texttt{VID} is
$$
T_{\texttt{VID}} = ``\text{airplane} ~ \text{antelope} ~ ... ~ \text{zebra}\text{''} \,,
$$
where each object category is converted to a candidate phrase to be grounded/aligned, parsed by blank spaces. 

As shown in Fig~\ref{fig: fw}, \texttt{TrIVD} encodes the above text prompt using a pre-trained transformer language model~\cite{Liu2019RoBERTaAR}, which produces a sequence of hidden vectors of same size as the input. In order to enforce the semantic alignment between the text and the visual (from the proposed object bounding boxes) representations, the projected feature embeddings of the text prompt are then supervised by: 

\noindent (1) Soft token loss ($\mathcal{L}_{\text{soft}}$);

\noindent (2) Region-text contrastive alignment loss ($\mathcal{L}_{\text{contrast}}$).


\paragraph{Soft Token Loss} The soft token prediction loss is a non-parametric alignment loss, which uses the positional information to align the proposed objects to text. As mentioned in Sec.~\ref{sec: vl}, instead of directly predicting a categorical class for each detected object with $N_{\text{cls}}$ fixed categories, for each proposed object, \texttt{TrIVD} outputs a token span which indicates the alignment scores between the original input text prompt and the proposed object.

Specifically, same as in MDETR~\cite{kamath2021mdetr}, we set the maximum number of tokens for any given sentence as $L = 256$. As a result, for each proposed object bounding box (query), \texttt{TrIVD} outputs a distribution uniformly over all token positions which correspond to the object~(Fig.~\ref{fig: fw}). For objects those are matched any ground truth target, the corresponding query is supervised by the non-object label $emptyset$. 

During training, the soft token loss is computed by cross-entropy between the predicted token span distribution probabilities and the actual token span of the ground truth target object category.


\paragraph{Contrastive Alignment Loss} The region-text contrastive alignment loss is a parametric loss which is used to encourage the similarity between the embedded representations of the proposed object queries (boxes) and the matched words of the input text prompt in the feature space, while dis-encouraging the closeness with those un-related words. Given the token spans' length as $L = 256$, and the number of proposed object queries (boxes) as $N_{\text{box}} = 500$. Denote the ground-truth tokens matched with $o_i$ target object as $T_i^+$, and the ground-truth objects that matched with a target token $t_i$ as $O_i^+$. The contrastive alignment loss reads:
\begin{equation}
    L_{\text{contrast}} = \frac{1}{2} (L_{\text{obj}} + L_{\text{tok}}) \,,
\end{equation}
where (1) $L_{\text{obj}}$ is the contrastive loss for all objects, which is normalized by number of positive tokens for each object ($\tau$ denotes the temperature and is set as 0.07 here),
\begin{equation}
    L_{\text{obj}} = \sum\limits_{i=0}^{N_{\text{box}}-1} \frac{1}{\vert T_i^+ \vert} \sum\limits_{j \in T_i^+} - log \bigg( \frac{exp(o_i^T t_j / \tau)}{\sum_{k=0}^{L-1} exp(o_i^T t_k / \tau) } \bigg) \,;
\end{equation}
And (2) $L_{\text{token}}$ is the contrastive loss for all tokens, normalized by the number of positive objects for each token,
\begin{equation}
    L_{\text{tok}} = \sum\limits_{i=0}^{L-1} \frac{1}{\vert O_i^+ \vert} \sum\limits_{j \in O_i^+} - log \bigg( \frac{exp(t_i^T o_j / \tau)}{\sum_{k=0}^{N_{\text{box}}-1} exp(t_i^T o_k / \tau) } \bigg) \,.
\end{equation}
\section{Additional Details on the Unified Detector-Tracker}
\label{app: detr}

As described in Sec.~\ref{sec: det_track}, \texttt{TrIVD} resorts to the self-attention mechanism and achieves both detection and tracking within one unified end-to-end model (Fig.~\ref{fig: fw}). Here, we provide more details on \texttt{TrIVD}'s \textbf{bipartite matching} procedure for the multi-task co-training of detection and tracking.

\paragraph{Unified Detection-Tracking Task} In a multi-task setting, one can view a detection task as a multi-class multi-object tracking task but with only one frame in the tracking video. I.e., all detected objects are simply regarded as \textit{newly appeared} objects. On the other hand, tracking requires data associations of the track identities across frames besides object localization and classification. 

\paragraph{Detection-Tracking Bipartite Matching}
\texttt{TrIVD} follows tracking-by-attention~\cite{meinhardt2021trackformer,Zhou2022GTR,Chu2021TransMOTSG} and realizes the data association via initializing the current frame's object queries with (1) empty object queries, for those newly appeared objects; and (2) previous detected object queries, i.e., track queries. 

Therefore, the resulting mapping $j = \pi(i)$ between the ground truth objects ($y_i$) to the predictions ($\hat{y_j}$) is determined by minimizing the overall costs under the following two scenarios: 

\begin{enumerate}

\item \label{det} (If the current task is detection.) The bipartite matching is determined via regular DETR-like detection costs based on bounding box similarity and object category~\cite{carion2020detr,zhu2020deformable}.

\item \label{track} (If the current task is tracking.) For tracking data association, each previously detected identities ($K_{t-1}$) will be kept and passed to the current frame ($K_{t}$). 
\begin{itemize}
\item[(a)] \label{exist} If the objects from previous frame also appear in the current frame ($K_{t-1} \cap K_{t}$), the bipartite matching is directly between their corresponding track identities; 

\item[(b)] \label{disappear} If the objects from previous frame disappear in the current frame ($K_{t-1} \setminus K_{t}$), the bipartite matching will assign to the background, i.e., non-object ($\emptyset$); 

\item[(c)] \label{new} If the objects are the newly appeared ones ($K_{t} \setminus K_{t-1}$), then the bipartite matching will be the same with the detection bipartite matching in Scenario \ref{det}.

\end{itemize}

\end{enumerate}

The matching of all above scenarios is achieved via searching the injective minimum cost mapping $\varphi$ in an assignment problem following DETR~\cite{carion2020detr,Stewart2015EndtoEndPD}. The resulting unified end-to-end training loss for \texttt{TrIVD} is therefore:
\begin{equation}
    \mathcal{L} = \mathcal{L}_{\text{soft}} + \mathcal{L}_{\text{contrast}} + \mathcal{L}_{\text{box\_detect}} + \mathcal{L}_{\text{box\_track}} \, ,
    \label{eq: loss}
\end{equation}
where $\mathcal{L}_{\text{soft}},\, \mathcal{L}_{\text{contrast}}$ are the object category prediction losses~(Sec.~\ref{sec: vl}, Appendix~\ref{app: classifier}), $\mathcal{L}_{\text{box\_detect}}$ refers to the matching scenarios \ref{det} and \ref{track}-(c). $\mathcal{L}_{\text{box\_track}}$ refers to the matching scenarios \ref{track}-(a-b).

Specifically, the box-based loss terms $\mathcal{L}_{\text{box\_detect}}$ and $\mathcal{L}_{\text{box\_track}}$ supervise bounding box differences by 1) $L_1$ distance and 2) the generalized intersection over union (IoU)~\cite{Rezatofighi2019GeneralizedIO} cost ($\mathcal{L}_{GIoU}$):
\begin{equation}
    \mathcal{L}_{\text{box}\_{\{\text{detect},\, \text{track} \}}} = \lambda_{L_1} \Vert b_i - \hat{b}_{\varphi(i)} \Vert_1 + \lambda_{GIoU} \mathcal{L}_{GIoU} (b_i, \hat{b}_{\varphi(i)}) \,,
\end{equation}
where $\lambda_{L_1}, \lambda_{GIoU} \in R^+$ are weights as hyperparameters.

\section{Additional Implementation Details}
\label{app: implement}

\subsection{Backbone Pre-training}
\label{app: backbone}

For the MViTv2-s~\cite{li2022mvit} backbone training, we follow the same recipe as in~\cite{liu2021Swin,fan2021mvit,li2022mvit}. Specifically, we pre-train MViTv2-s on \texttt{ImageNet-21K}~\cite{Deng2009ImageNetAL} for 300 epochs with batch size of 32. We use the truncated normal distribution initialization~\cite{Hanin2018HowTS} and synchronized AdamW~\cite{Loshchilov2017FixingWD} optimization, with a base learning rate of $2 \times 10^{-3}$ and a linear warm-up in the first 70 epochs followed by a decayed half-period cosine schedule~\cite{Touvron2021TrainingDI}. We set the weight decay to 0.05. We also use stochastic depth~\cite{Huang2016DeepNW} with rate as 0.1. The augmentation strategies are the same as in ~\cite{fan2021mvit,li2022mvit}.

\subsection{Text Encoder}
\label{app: text}

We follow~\cite{kamath2021mdetr} and use the HuggingFace~\cite{Wolf2020TransformersSN} pre-trained RoBERTa-base~\cite{Liu2019RoBERTaAR} as our text encoder. In all the experiments, we use a linear decay with warm-up schedule, increasing linearly to $5\times10^{-5}$ during the first 1\% of the total number of iterations, then decreasing linearly back to 0 for the rest of the training.


\subsection{Single-dataset Training}
\label{app: single}

For the single-dataset training of \texttt{TrIVD$_{\text{backbone}}$}, we fine-tune the pre-trained model from Appendix~\ref{app: backbone} on the downstream object detection and multi-object tracking tasks respectively.


\paragraph{COCO Image Object Detection}
\label{app: coco}
For image object detection experiments on \texttt{COCO}~\cite{lin2014coco}, we follow the 3$\times$schedule (36 epochs) suggested in~\cite{li2022mvit}. We set the batch size as 2 with an initial learning rate of $10^{-4}$ for deformable DETR encoder-decoder, and $10^{-5}$ for the backbone. The learning schedule for the text encoder is detailed in Appendix~\ref{app: text}.

\paragraph{VID Video Object Detection}
\label{app: vid}
For video object detection experiments on \texttt{VID}~\cite{russakovsky2015vid}, we follow the previous work~\cite{Deng2019RelationDN,Wang2018FullyMN,zhu2017dff,Yao2020VideoOD} and include \texttt{DET} ~\cite{Russakovsky2015ilsvrc} dataset in the pre-training stage. Then, we train our model on \texttt{VID}~\cite{russakovsky2015vid} with batch size as 1 for 3 epochs. The initial learning rate for deformable DETR is $5\times 10^{-4}$, and $5\times10^{-5}$ for the backbone. The learning schedule for the text encoder is detailed in Appendix~\ref{app: text}.

\paragraph{MOT17 Multi-object Tracking}
\label{app: mot}
For multi-object tracking experiments on \texttt{MOT17}~\cite{Milan2016MOT16AB}, we initialize the model weights from the model trained on \texttt{COCO}~\cite{lin2014coco} as described in Appendix~\ref{app: coco}. We train our model on \texttt{MOT17}~\cite{Milan2016MOT16AB} with a batch size of 2 for 50 epochs with a learning rate drop to $0.1\times$ after 30 epochs. The initial learning rate for deformable DETR is $5\times 10^{-4}$, and $5\times10^{-5}$ for the backbone. The learning schedule for the text encoder is detailed in Appendix~\ref{app: text}.

\subsection{Cross-dataset Co-training}
\label{app: multi}

For the joint training of \texttt{TrIVD$_\text{multitask}$} on all the three datasets (\texttt{COCO}, \texttt{VID},  \texttt{MOT}), we initialize the model weights from the model trained on \texttt{COCO}~\cite{lin2014coco} as in Appendix~\ref{app: coco}. We then co-train our model on the combined dataset of \texttt{COCO}, \texttt{VID} and \texttt{MOT} with a batch size of 1 for 120 epochs with a learning rate drop to $\times 0.1$ after 80 epochs. The initial learning rate for deformable DETR is $10^{-4}$, and $10^{-5}$ for the backbone. The learning schedule for the text encoder is detailed in Appendix~\ref{app: text}. 

To handle the different tasks (image object detection, video object detection, or multi-object tracking) for samples from different datasets, we separate samples from the three datasets in each forward pass. To balance the varying scales of different datasets, we randomly select 20,000 video clip samples from \texttt{VID}~\cite{russakovsky2015vid} in each training epoch. See further discussions on balancing detection and tracking data during \texttt{TrIVD}'s multi-task co-training in Appendix~\ref{app: data_balance}.


\subsection{Track Initialization and Re-identification}
\label{app: reid}

New objects appearing in the current frame compared to previous frames are detected by a fixed number of $N_{\text{box}} = 500$ object queries~\cite{zhu2020deformable}, each attends to certain spatial locations in the current frame~\cite{zhu2020deformable}. Given the object encodings, the deformable-DETR transformer decoder's self-attention intrinsically avoids duplicate detections~\cite{carion2020detr,zhu2020deformable,meinhardt2021trackformer}. For the tracking purpose, each newly detected objects will also initialize a new track query with its associated object embedding~\cite{meinhardt2021trackformer}. Track queries then follow the corresponding objects based on their embeddings throughout a video and adapt to the position changes simultaneously. Depends on the objects status in the given video sequences, the number of track queries $N_{\text{track}}$ could change across frames each time when new objects are detected or previously-detected tracks disappear or are occluded. Following~\cite{meinhardt2021trackformer}, we remove a detection/track when its classification confidence score drops below $\sigma_{\text{track}}=0.4$, or is lower than an IoU threshold $\sigma_{\text{NMS}}=0.9$ for non-maximum suppression (NMS).

During tracking inference, we use previously proposed track queries for an attention-based re-identification process. We follow~\cite{meinhardt2021trackformer} and keep previously removed track queries within an optimal inactive patience of $N_{\text{reid}}=5$ frames, during which the track queries are considered as not active and thus are not used in the object queries initialization of new frames, unless a classification score higher than $\sigma_{\text{reid}}=0.4$ triggers the re-identification.

\subsection{Track Filtering}
\label{app: filter}

Typical tracking-by-detection methods~\cite{Kim2015MultipleHT,LealTaix2016LearningBT,Henschel2017ImprovementsTF,Chen2018RealTimeMP,Keuper2020MotionS,LealTaix2011EverybodyNS} perform data association on a bounding box level during tracking evaluations. Yet this strategy is not suitable for tracking-by-attention or point-based methods~\cite{Zhou2020TrackingOA,meinhardt2021trackformer}. To achieve a fairer comparison, we follow~\cite{meinhardt2021trackformer} and perform the track filtering on Intersection over Union (IoU) and initialize tracks with IoU greater than 0.5.

\subsection{Evaluation Metrics for Multi-object Tracking}
\label{app: eval}
We provide brief definitions of the seven evaluation metrics used for MOT comparisons~(Sec.~\ref{sec: compare})
in the main paper. For more complete analyses on different metrics for multi-object tracking, please also refer to ~\cite{Ristani2016PerformanceMA,Milan2016MOT16AB}.

\paragraph{False Negative (FN)} refers to the number of false negative ground truth bounding boxes that are not covered by any bounding box.

\paragraph{False Positive (FP)} refers to the number of false positive bounding boxes that do not correspond to any ground truth object.

\paragraph{Multiple Object Tracking Accuracy (MOTA)}penalizes detection
errors (FN $+$ FP) and fragmentations ($\Phi$) normalized by the total number $N$ of true detections:
\begin{equation}
\text{MOTA} = 1 - \frac{\text{FN} + \text{FP} + \Phi}{N}\,.
\end{equation}

\paragraph{IDF1} is defined as the ratio between the correctly identified detections and the average number of ground truth objects and computed detections:
\begin{equation}
\text{IDF1} = \frac{2\, \text{IDTP}}{2\,\text{IDTP} + \text{IDFP} + \text{IDFN}}\,,
\end{equation}
where IDP refers to the identification precision. And IDTP, IDFP, IDFN are the true positive, false positive, and false negative of IDP respectively.

\paragraph{Mostly
Tracked (MT)} denotes the number of tracks that are successfully tracked for larger than 80\% of its total span.

\paragraph{Mostly
Lost (ML)}denotes the number of tracks that are successfully tracked for less than 20\% of its total length.

\paragraph{Identity Switch (IDS)} counts the number of mismatches of a ground truth object that is originally identified as track $i$ but assigned to another track $j$ ($i \neq j$) in the following frames.

\section{Further Discussions on \texttt{TrIVD}}
\label{app: discussion}


\subsection{Zero-shot Tracking Vocabulary}
\label{app: zero-shot}

\texttt{TrIVD} learns to detect from all the \texttt{COCO} $\&$ \texttt{VID} $\&$ \texttt{MOT17} data but learns to track only from the person-tracking data (\texttt{MOT17}), yet it is able to track all categories \textit{as long as} they are sufficiently provided in detection training data. This can't be achieved without our unified detector-tracker design. Practically, tracking datasets are usually small-sized and of limited categories, yet \texttt{TrIVD} can track objects by learning from their detection data without requiring their tracking ground truth.

For instance, we assume the vocabulary (annotated object categories) of a collection of detection (image and/or video) datasets, $\mathcal{D}_{\text{det}} = \{\mathcal{D}^i_{\text{det}}, \, i = 0,\, \dots ,\, N_{\text{det}}\}$, is $\mathcal{V}_{\text{det}}$, and the vocabulary of a collection of multi-object tracking datasets, $\mathcal{D}_{\text{mot}} = \{\mathcal{D}^i_{\text{mot}}, \, i = 0,\, \dots ,\, N_{\text{mot}}\}$, is $\mathcal{V}_{\text{mot}}$. Then a well-trained \texttt{TrIVD} on the combined dataset of $\mathcal{D}_{\text{det}} + \mathcal{D}_{\text{mot}}$ should have the detection/tracking vocabulary as 
\begin{equation}
    \mathcal{V}_{\text{TrIVD}}\big\vert_{\text{detect}} \equiv \mathcal{V}_{\text{TrIVD}}\big\vert_{\text{track}} := \mathcal{V}_{\text{det}} \cup \mathcal{V}_{\text{mot}} \,.
\end{equation}
More specifically, we list the following examples to help readers better understand what \texttt{TrIVD} \textit{can} or \textit{cannot} detect or track (color ``blue'' denotes regular tracking, color ``red'' denotes zero-shot tracking): 

\begin{table}[h]
\centering
\resizebox{\linewidth}{!}{  
    \begin{tabular}{ccccc}
       \toprule
      \textbf{Object Example} & {\bf $\mathcal{V}_{\text{det}}$} & {\bf $\mathcal{V}_{\text{mot}}$} & \textbf{Detect} & \textbf{Track} \\ 
      \hline 
     person & $\in$ & $\in$ & \cmark & {\color{blue}\cmark} \\ 
     car & $\notin$ & $\in$ & \cmark & {\color{blue}\cmark} \\ 
     panda & $\in$ & $\notin$ & \cmark & {\color{red}\cmark} \\ 
     vinyl & $\notin$ & $\notin$ & \xmark & {\color{red}\xmark} \\ 
\bottomrule  
    \end{tabular}  
}
\end{table}

As listed in the above table, benefited from the detection-tracking co-training, \texttt{TrIVD} extends the tracking vocabulary to the combined vocabulary of available detection and tracking annotated data ($\mathcal{V}_{\text{det}} \cup \mathcal{V}_{\text{mot}}$), and thus achieves zero-shot tracking. 

However, for the case of ``vinyl'', which belongs to an object category not included in $\mathcal{D}_{\text{det}}$ nor $\mathcal{D}_{\text{mot}}$, \texttt{TrIVD} will not be able to detect or track it. To further expand \texttt{TrIVD}'s tracking vocabulary, one can pre-train it on large-scale, and semantic-rich detection or phrase grounding datasets such as Objects365~\cite{Shao2019Objects365AL}, GLIP~\cite{li2021glip,Zhang2022GLIPv2UL}, Flickr30K~\cite{Plummer2015Flickr30kEC}, VG Caption~\cite{Krishna2016VisualGC}, etc.


\subsection{How Should the Multi-task
Data be Organized during Co-training?}
\label{app: data_balance}

From our experiments, the amount of data for a normal detector to detect well equals those for \texttt{TrIVD} to detect/track well:

\begin{enumerate}
    
\item As a unified detector/tracker, \texttt{TrIVD} is a detector in the first place. This means the number of required training samples of a category for a normal detector to perform well, is the \textit{same} as what \texttt{TrIVD} needs to detect well. 

\item On top of a detector, \texttt{TrIVD} is also a multi-object 
 tracker, and faces the same challenges as regular trackers. Given sufficient tracking training samples (which could be from different categories) for a normal tracker to perform well, \texttt{TrIVD} is able to track all the categories it can detect (See explanations in Appendix~\ref{app: zero-shot}). 

\item Therefore, in \texttt{TrIVD}$_{\text{multitask}}$ setup, we first pre-train \texttt{TrIVD} on the category-rich \texttt{COCO} image OD dataset, then co-train \texttt{TrIVD} across 3 datasets with \textit{same} sampling size (as described in Appendix~\ref{app: multi}). Specifically, we found an increase in detection performance for ``person'' class after adding \texttt{MOT17} person-tracking data (from 68.5 to 72.3 AP), which is consistent with the co-training intuition that more data leads to better performance.

\end{enumerate}


\subsection{Multi-task Co-training's Mutual Benefits between Detection and Tracking}
\label{app: detect-track}

\paragraph{Detection Benefits Tracking}
The benefits from detection to tracking in \texttt{TrIVD}'s multi-task co-training framework is obvious and significant. As discussed in Appendix~\ref{app: zero-shot}, the detection-tracking co-training setup enables \texttt{TrIVD} to extend its tracking vocabulary to the \textit{combined} vocabulary of available annotated detection and tracking data, and thus achieves zero-shot tracking and largely scales up the tracking categories. 

\paragraph{Tracking Benefits Detection} As also revealed by recent open-vocab OD work~\cite{jiang2019lwdn,LealTaix2016LearningBT,zang2022ovdetr}, \textit{larger datasets \& richer categories} are keys to boosting OD performance. As tracking is formatted as detection plus data association in our unified setup, tracking data helps \texttt{TrIVD} to learn detection with its additional data and categories: \\

\resizebox{\linewidth}{!}{
\hspace*{-0.3cm}
    \begin{tabular}{ccccc}
       \toprule
      \multicolumn{1}{c}{\multirow{2}{*}{\textbf{Object}}} &  \multicolumn{1}{c}{\multirow{2}{*}{\textbf{\texttt{COCO}~\cite{lin2014coco}}}} & \multicolumn{1}{c}{\multirow{2}{*}{\textbf{\texttt{MOT17}~\cite{Milan2016MOT16AB}}}} & \multicolumn{2}{c}{\multirow{1}{*}{\textbf{Detect AP ($\uparrow$)}}} \\ [-0.6ex]
        
        \cmidrule(lr){4-5} 
     & & & {\textbf{\texttt{TrIVD}$_{\text{backbone}}$}} & {\textbf{\texttt{TrIVD}$_{\text{mutlitask}}$}} \\ 
     \midrule 
      person  & \cmark & \cmark & 75.8 & 79.3 \\  
      non-person & \cmark & \xmark & 35.7 & 36.0  \\ 
\bottomrule  
    \end{tabular} 
}

\begin{enumerate}

\item As shown in Tabs.~\ref{tab: coco}-\ref{tab: vid}, for both image OD and video OD tasks, \texttt{TrIVD}$_{\text{multitask}}$ outperforms \texttt{TrIVD}$_{\text{backbone}}$. And the gain comes \textit{entirely} from the unified design and detection-tracking co-training strategy; 

\item More specifically, as listed in the above table, when examining the object category-level detection performance, we observe an increase of detection AP in \texttt{COCO}'s ``person'' class from \texttt{TrIVD}$_{\text{backbone}}$ (75.8) to \texttt{TrIVD}$_{\text{multitask}}$ (79.3). With same model structure between \texttt{TrIVD}$_{\text{backbone}}$ and \texttt{TrIVD}$_{\text{multitask}}$, the gain of 3.5 in AP is entirely from the extra ``person'' data in \texttt{MOT17} person-tracking dataset during the multi-task co-training of \texttt{TrIVD}$_{\text{multitask}}$ on \texttt{COCO} \& \texttt{MOT17}. Meanwhile, we have not seen significant improvement regarding the non-person category after \texttt{TrIVD}'s multi-task co-training strategy, since \texttt{MOT17} does not contain annotated non-person objects and therefore extra training data from \texttt{MOT17} will not help the detection performance on non-person objects.

\end{enumerate}


\subsection{\texttt{TrIVD} as a General Detection-Tracking Framework for Further Improvements}
\label{app: general}

\texttt{TrIVD} presents a \textit{general} unified framework integrating self-attention for detection and tracking, whose detection component is modified from Deformable-DETR~\cite{zhu2020deformable}, and tracking-by-attention mechanism is modified from TrackFormer~\cite{meinhardt2021trackformer}. Therefore in Sec.~\ref{sec: compare}, we show \texttt{TrIVD}'s better performance versus our detector/tracker baselines, to demonstrate that unified detection-tracking multi-task co-training strategy benefits both detectors and trackers. 

Essentially, \texttt{TrIVD} can be \textit{adapted to any} transformer-based detector and tracker. E.g., upgrading 
our transformer decoder following newer detection or tracking methods will equip \texttt{TrIVD} with more advanced detector/tracker designs and therefore further improve its overall detection-tracking performance. To name a few, switching \texttt{TrIVD}'s transformer decoder to the spatial graph decoder of TransMOT~\cite{Chu2021TransMOTSG}, or the global tracking decoder of GTR~\cite{Zhou2022GTR}, will equip \texttt{TrIVD} with more advanced tracker designs and thus boost its tracking performance.

\subsection{Classic Cross-entropy v.s. Region-text Alignment for Class Prediction}
\label{app: cross-entropy}

They have similar performances but visual-text classifier is more general and flexible. We in fact explored multi-class CE loss first, and switched to the current visual-text classifier: 

\begin{enumerate}

\item More flexibility. CE loss based on logit-encoded labels is not generalizable for multi-dataset co-training -- One needs to re-arrange/add class labels each time a new dataset is added to the co-training. Based on semantic meanings and visual-text alignment, our grounding-based classifier does not need to be modified and is more \textit{flexible} when we pre-train \texttt{TrIVD} on large-scale datasets or later fine-tune it on additional datasets for other downstream tasks. 

\item Same performance. We found similar classification/detection scores are achieved by visual-text alignment (46.5 AP in \texttt{COCO}, 78.3 AP in \texttt{VID}) and CE loss (46.3 AP in \texttt{COCO}, 78.4 AP in \texttt{VID}).

\end{enumerate}



\end{document}